\documentclass[runningheads]{llncs}
\usepackage[T1]{fontenc}
\usepackage{graphicx}
\usepackage{times}

\usepackage{multicol}
\usepackage{changes}
\usepackage[bookmarks=true]{hyperref}

\usepackage{amsmath}
\usepackage{mathtools}
\usepackage{textcomp}
\usepackage{algorithmic}
\usepackage[ruled,vlined,noend,linesnumbered]{algorithm2e}
\usepackage[english]{babel}
\usepackage{amssymb}
\usepackage{graphicx}
\usepackage{xcolor}
\usepackage{enumerate}
\usepackage{framed}
\usepackage{tabularx}
\usepackage{url}
\usepackage{mathrsfs}
\usepackage[framemethod=default,innerleftmargin=5pt,innerrightmargin=5pt,skipabove=5pt,topline=false,rightline=false,bottomline=false,linewidth=2.5pt,roundcorner=5pt]{mdframed}
\usepackage{paralist}

\usepackage{xcolor}
\usepackage[linesnumbered,ruled,vlined]{algorithm2e}

\SetCommentSty{mycommfont}

\SetKwInput{KwInput}{Input}                
\SetKwInput{KwOutput}{Output}  

\usepackage{complexity}
\let\oldNP\NP
\renewcommand{\NP}{\oldNP\xspace}

\newcommand{\pagebudget}[1]{}
\newcommand{\showtotalpagebudget}[1]{}

 \newtheorem{assumptionenv}{Assumption}

 \newenvironment{assumption}{\vspace{0.05in}\begin{assumptionenv}\em}{\end{assumptionenv}\vspace{0.05in}}

	\newcommand{\etal}{{et~al.}}

 \newcommand*{\probleminternal}[4]{
	\par
	\medskip
	\noindent\fbox{\parbox{0.98\columnwidth}{
			\textbf{#4: #1} \\[0.05in]
			\renewcommand{\tabcolsep}{2pt}
			\begin{tabularx}{\linewidth}{X}
				\textbf{Input:} #2 \\
				\textbf{Output:} #3
			\end{tabularx}
		}}
		\par
		\medskip
		\par
	}

	\newcommand*{\problemdef}[3]{\probleminternal{#1}{#2}{#3}{Problem}}
	\newcommand*{\decproblem}[3]{\probleminternal{#1}{#2}{#3}{Decision Problem}}

	\makeatletter
	\newcommand*{\Relbarfill@}{\arrowfill@\Relbar\Relbar\Relbar}
	\newcommand*{\xeq}[2][]{\ext@arrow 0055\Relbarfill@{#1}{#2}}
	    \makeatother

\newcommand{\removed}[1]{}

\newcommand{\walks}{\operatorname{Walks}}
\newcommand{\src}{{\rm src}}
\newcommand{\srcfunc}[1]{\src({#1})}
\newcommand{\tgt}{{\rm tgt}}
\newcommand{\tgtfunc}[1]{\tgt({#1})}
\newcommand{\pow}[1]{\ensuremath{\raisebox{.15\baselineskip}{\Large\ensuremath{\wp}}({#1})}\xspace}

\newcommand{\twin}{{\rm twin}}
\newcommand{\edge}{{\rm edge}}
\newcommand{\red}{{\rm red}}
\newcommand{\blue}{{\rm blue}}
\newcommand{\gray}{{\rm gray}}

\newcommand{\Yes}{{\rm Yes}\xspace}
\newcommand{\No}{{\rm No}\xspace}
\newcommand{\MCSD}{{\rm MCSD}\xspace}
\newcommand{\MCSDDEC}{{\rm MCSD-DEC}\xspace}
\newcommand{\DMCDEC}{{\rm DMC-DEC}\xspace}
\newcommand{\CDMCDEC}{{\rm CDMC-DEC}\xspace}
\newcommand*{\gobble}[1]{}

\begin{document}
\title{Optimal Sensor Deception to Deviate from an Allowed Itinerary}
%
%
\author{Hazhar Rahmani\inst{1}\orcidID{0000-0002-6342-2273} \and
Arash Ahadi\inst{2}\orcidID{0009-0008-9617-1301}  
\and\\
Jie Fu\inst{3}\orcidID{0000-0002-4470-2827}}
\authorrunning{H. Rahmani et al.}


%
\institute{Missouri State University, Springfield MO 65897, USA \and
University of Windsor, Windsor ON N9B 3P4, Canada \and University of Florida, Gainesville FL 32601, USA\\
\email{hrahmani@missouristate.edu},
\email{fujie@ufl.edu}, 
\email{arash.ahadi@uwindsor.ca} \\
}

%
\maketitle              
\begin{abstract}
In this work, we study a class of deception planning problems in which an agent aims to alter a security monitoring system's sensor readings to disguise its adversarial itinerary as an allowed itinerary in the environment. Both the adversarial and allowed itinerary sets are defined by regular languages. 
We investigate whether there exists a strategy for the agent to alter the sensor readings, with a minimal cost, such that for any adversarial paths it takes, the system thinks the agent took a path within the allowed itinerary. Our formulation assumes an offline sensor alteration where the agent determines the sensor alteration strategy and implements it, and then carries out any path in its adversarial itinerary. We prove that the problem of solving the optimal sensor alteration is NP-hard and present an exact algorithm based on integer linear programming. We demonstrate the correctness and the efficacy of the algorithm using several experiments.
\end{abstract}
\vspace{-24pt}
\section{Introduction}
%

Sensors are crucial in robotics and intelligent systems for gathering data.
%
In some applications, the history of sensor readings is used for monitoring or controlling the spatio-temporal activities of agents within the environment.
Consider the \emph{story validation problem} proposed by Yu and LaValle~\cite{yu2010cyber,yu2011story}, in which an agent claims that it has visited a sequence of regions within an environment.
The system's task is then to check, using the history of sensor readings, whether the agent claim is invalid.
%
%
%
Rahmani \emph{et al.} ~\cite{rahmani2021sensor} studied sensor design for the story validation problem, casting it as a \emph{sensor selection problem}, and developed an algorithm to choose a  minimal number of sensors, if they exist, so that, based on the sensor readings, the system is able to tell whether the agent followed its itinerary. 
%
%
Phatak and Shell~\cite{phatak2023sensor} extended this sensor selection problem to include multiple itineraries, which serve as behavioral patterns. The aim was to determine, for any execution, to which of those patterns the execution matches.
%
%
%
%
%

The story validation and sensor selection problems have practical applications in surveillance, security patrolling, and activity tracking.  However, existing
solutions  do not consider the case where a deceptive attacker exploits the sensing system's partial observations to hide its adversarial intention.
Addressing the vulnerabilities of a sensor network design for those applications is  crucial for the security of the system. 
%

%
  

\begin{figure}[t]
  \centering
  \includegraphics[width=0.7\linewidth]{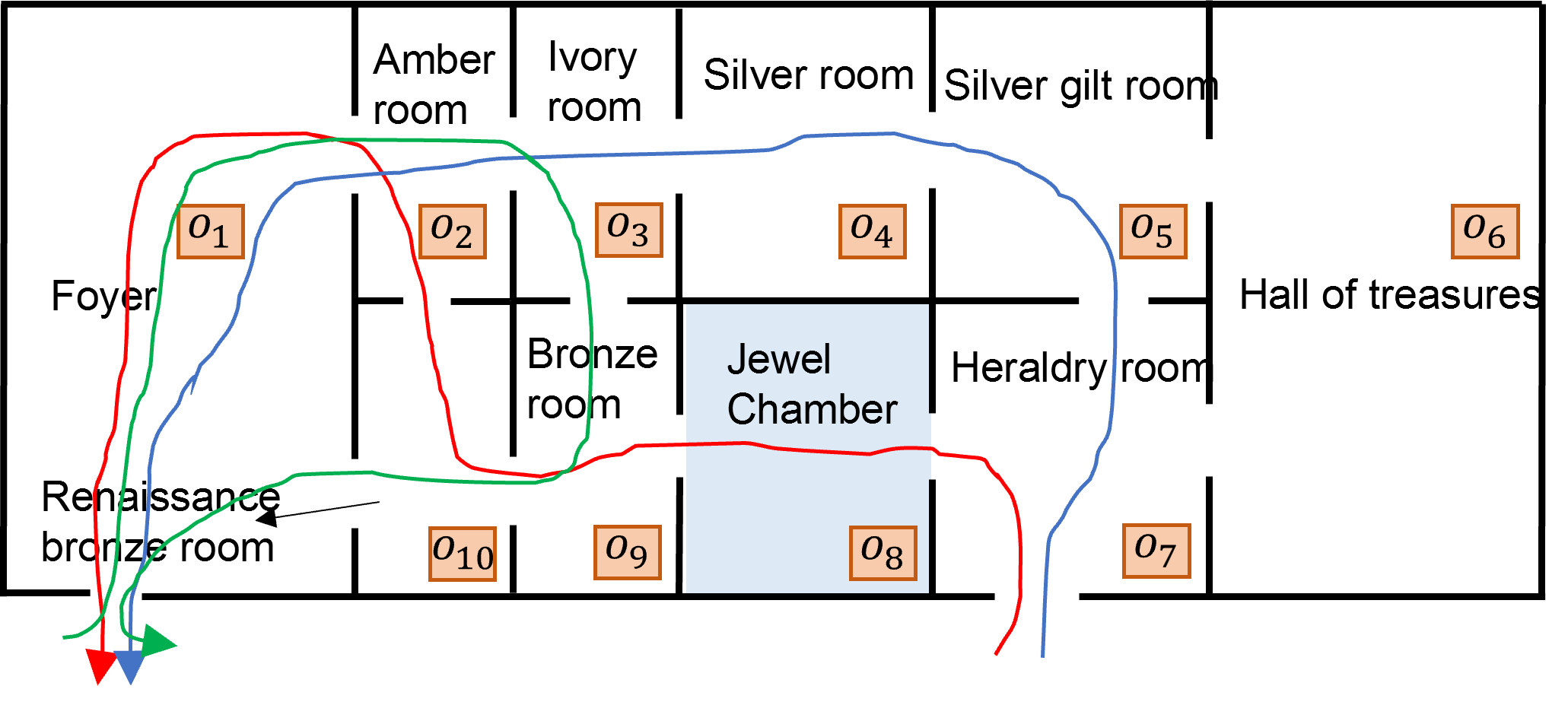}
  \vspace*{-18pt}

  \caption{
    The floor map of Green Vault. The blue and green routes are allowed. The red route is the adversary's itinerary. $o_i$'s are occupancy sensors. 
  \label{fig:greenVault}
  }
\vspace*{-18pt}
\end{figure}

In this paper, we study the following problem: An agent intends to take a tour in the environment and it is allowed to take specific paths.  
If the agent can carry out attacks to alter sensor readings, does there exist a sensor alteration strategy enabling any deviated path (attack intention) to appear observation-equivalent to an allowed path? In addition, given the costs of the sensor attacks, how to compute a minimal sensor alteration strategy to achieve the deception goal? By answering these questions, we can identify vulnerabilities of a security sensor network and assess the level of security by the cost associated with compromising the security.
To motivate, consider Fig.~\ref{fig:greenVault}, displaying the floor map of the Green Vault museum in Germany, which suffered a \$123 million jewel heist in 2019. 
Each room is guarded by an occupancy sensor $o_i$, triggering an \emph{activation} event when an agent enters that room and producing a \emph{deactivation} event when the agent leaves the room.
%
%
The agent is allowed to take any of the two tours in blue and green. 
When the agent finishes its tour of the environment, the system can verify using the events produced by the sensors, whether the agent took any of those two allowed tours.
The agent, with adversarial intent, aims to take the tour shown by the red route.
%
In this specific example, it is possible to do so using the sensor swaps $o_5 \leftrightarrow o_8$, $o_4 \leftrightarrow o_9$, $o_3 \leftrightarrow o_{10}$. The cost of this sensor alteration attack is $6$, $2$ for each swapping.

The formulated problem is a class of intention deception problems using sensor alteration attacks. Karabag \etal~\cite{karabagDeceptionSupervisoryControl2021a} studied a probabilistic planning approach where the optimal plan minimizes the Kullback–Leibler (KL) divergence between the (observations of) agent's strategy and (that of) the reference strategy, provided by the supervisor. 
Fu \cite{fuAlmostSureIntentionDeception2022} studied the intention deception planning problem in which an attacker deceives the system into a wrong belief about the agent's intention, by ensuring the trajectory that satisfies the adversary's intention is observation-equivalent to a trajectory that satisfies an allowed intention.  Masters and Sardina \cite{mastersDeceptivePathplanning2017} studied a plan obfuscation problem where an agent plans a path to reach its goal state while making  an observer unable to recognize its goal until the last moment. In aforementioned work, the intention deception exploits the noises in the dynamics and/or imperfect observations but does not employ sensor attacks. On the other hand,
sensor deception has  been extensively studied in supervisory control  \cite{meira2020synthesis,meira2019synthesis,meira2021synthesis,mohajerani2020efficient,yao2022sensor,zheng2021modeling,wangSupervisoryControlDiscrete2019}. 
 Meira-Góes \emph{et al.}~\cite{meira2020synthesis,meira2019synthesis,meira2021synthesis} consider sensor deception for discrete systems where the attacker inserts, deletes, or edits sensor readings to induce the supervisor into allowing the system to enter an unsafe state. 
 They also proposed a method to synthesize a supervisor preventing the system from reaching unsafe states despite sensor deception attacks~\cite{meira2021synthesissupervisors}.
 %
%
Zheng\emph{et al.}~\cite{zheng2021modeling} consider joint sensor-actuator attacks where the attacker can simultaneously attack some sensors and actuators to deviate the designed behavior of the system.
%
%
Wang and Pajic~\cite{wangSupervisoryControlDiscrete2019} consider supervisory control under attacks on sensors and actuators where the attacks are modeled by finite state transducers. The main differences between sensor attack in supervisory control and our work are 1) the attacker is also the controlled agent who carries out the deviated itinerary; and 2)  we consider an offline attack in which the attack strategy does not depend on the history of observations. This offline attack strategy must ensure \emph{no matter which deviated itinerary is taken after the attack, the defender won't be able to detect the deviation. } 
This attack model is motivated by situations, such as the heist, where the attack is performed on the physical
infrastructure/devices, and where the attacker swaps the locations of the sensors in the environment.

 To this end, we model the agent dynamics using a discrete deterministic transition system called a world graph. An allowed itinerary is given as words accepted by an itinerary automaton. The adversary's intention is specified by a deviation automaton, describing all deviation paths. The world is equipped with a finite set of sensors and an observation function for the system. The sensors are susceptible to replacement attacks: The agent can replace the sensor A's reading by B's reading. 
 A sensor alteration is deceptive if any path in the deviation itinerary is made observation-equivalent to a path in the allowed itinerary, when the system receives the altered sensor readings.
 
 After presenting preliminaries and problem formulation in Section~\ref{sec:defn}, we prove the $\NP$-hardness of the problem in Section~\ref{sec:hard}. Then, in Section~\ref{sec:ILP} we employ an automata-theoretic approach to formulate an integer linear program for computing an optimal solution. Finally, we demonstrate the performance of our algorithm in Section~\ref{sec:case}.

\vspace{-0.5\baselineskip}
\section{Preliminaries and Problem Formulation}\pagebudget{1.7}
\label{sec:defn}
\subsection{Modeling the environment}
        We model the environment using the following structure.
        \begin{definition}[World graph]\label{def:wg}
		A \emph{world graph} is an edge-labeled directed multigraph $\mathcal{G} = (V, E, \src, \tgt, v_0, S,\mathbb{Y},\mathcal{O})$ in which 
		\begin{inparaenum}[(1)]
		    \item $V$ is the set of vertices, 
		    \item $E$ is the set of directed edges,
		    \item $\src: E \rightarrow V$ is the \emph{source} function, which identifies the source vertex of each edge,
		    \item $\tgt: E \rightarrow V$ is the \emph{target} function, which identifies the target vertex of each edge, 
		    \item $v_0 \in V$ is the initial vertex, 
		    \item $S = \{s_1, s_2, \dots, s_k\}$ is a nonempty finite set of sensors, 
		    \item $\mathbb{Y} = \{Y_1, Y_2, \dots, Y_k\}$ is a collection of mutually disjoint event sets where for each $i \in \{1, 2, \dots, k \}$, $Y_i$ is the events associated with $s_i$, and we use $Y$ to denote $Y_1 \cup Y_2 \dots Y_k$, i.e., $Y = Y_1 \cup Y_2 \dots Y_k$, and
		    \item $\mathcal{O}: E \rightarrow  \pow{Y} $ is a \emph{observation function}, which assigns to each edge, a \emph{world-observation}, which is a set of events that happen simultaneously when the agent takes the transition corresponding to that edge. We assume $\mathcal{O}(e) \neq \emptyset$ for all $e \in E$. (Here $\pow{X}$ denotes the set of all subsets of $X$.)
		\end{inparaenum}
		\end{definition}
        %
        %
        %
        
        A world graph describes the environment's topology and it identifies the sensors in the environment, along with the sensor events triggered by the agent motion.
        Each vertex of the graph is a region in the environment, and each edge indicates a feasible transition between two regions. The label of an edge identifies sensor events that are triggered simultaneously when the agent takes the transition corresponding to that edge. Those labels are indicated by the observation function $\mathcal{O}$.
        A sensor $s_i$ produces a set of events $Y_{i}$. 
        Definition~\ref{def:wg} assumes the event sets produced by distinct sensors are disjoint, that is, for any tuple of sensors $s_i, s_j \in S$ such that $i \neq j$, $Y_{i} \cap Y_{j} = \emptyset$.
        %

\begin{figure}[t]
  \centering
  \includegraphics[width=1.0\linewidth]{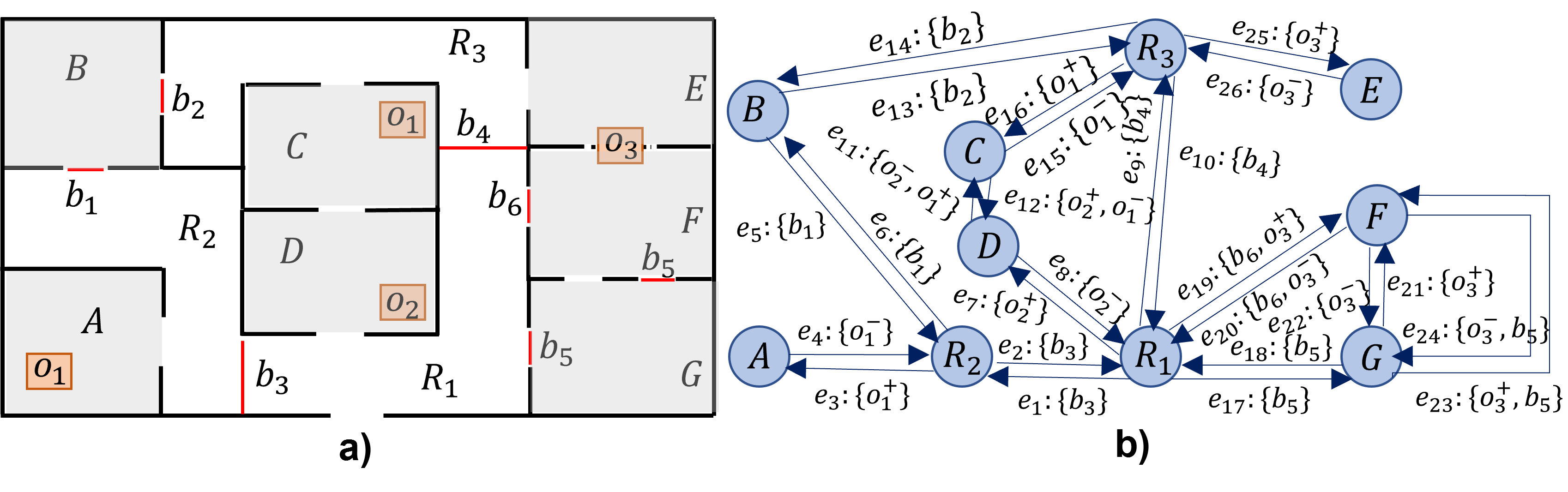}
  \vspace*{-2\baselineskip}
\caption{
    \textbf{a)} 
    A department guarded by beam sensors $b_1$ through $b_5$ and occupancy sensors $o_1$, $o_2$, and $o_3$.
    %
    \textbf{b)} A world graph modeling the department's map along with the located sensors. 
  \label{fig:department}}
\vspace*{-18pt}
\end{figure}

       %
        To illustrate, we provide an example with only beam and occupancy sensors.
       %
       %
        %
\begin{example}
    The environment in Fig.~\ref{fig:department}a, guarded by beam sensors $b_1$ through $b_5$ and occupancy sensors $o_1$, $o_2$, and $o_3$, is modeled using the world graph in Fig.~\ref{fig:department}b.
A beam sensor detects the passage of an agent between two adjacent regions, without detecting the direction of the passage.
Accordingly, to each beam sensor $s_i = b$, an event $b$ is assigned, i.e., $Y_{i} = \{b\}$, which occurs only when the agent crosses $s_i$.
An occupancy sensor $o$ detects the presence of an agent within a region.
When the agent enters that region, $o$ is activated, and when the agent leaves that region, $o$ is deactivated.
As a result, to sensor $s_i = o$, two events, $o^+$ and $o^-$, which respectively denote the activation and the deactivation of the sensor, are assigned, i.e., $Y_{i} = \{o^+, o^-\}$.
Note that if the agent does not enter the region, $o$ is not triggered and, thus, $o$ does not produce any event.
%

%

Note that because the world graph is a multi-graph, it allows to have multiple transitions between two regions.
For instance, there are two doors between $F$ and $G$, and thus, in the world graph there are two transitions from $F$ to $G$, and vice versa.
%
Also, a sensor can be used to guard multiple regions.
Sensor $o_1$ guards both rooms $A$ and $C$. 
Room $G$ has three doors, two of which are guarded by $b_5$.
Sensor $o_3$, located on the window between rooms $E$ and $F$, guards both $E$ and $F$. 
Because $E$ and $F$ are separated by a window (rather than a door), an agent cannot directly transit between them.
\end{example}
%
%
%
%
%
A tour in the environment is represented by a \emph{walk}, a sequence of edge $e_1 e_2 \cdots e_n \in E^*$ on the world graph $\mathcal{G}$ where
$\srcfunc{e_1}=v_0$ and $\tgtfunc{e_i}=\srcfunc{e_{i+1}}$ for each $i \in \{ 1, 2, \ldots, n-1\}$.
%
The set of all walks over $\mathcal{G}$ is denoted $\walks(\mathcal{G})$.
Note that a walk does not necessarily need to be a simple path in the world graph.

To specify a set of tours, we use a \emph{nondeterministic finite automaton} (NFA) over $E$.
\vspace{-12pt}
\begin{definition}
\label{def:NFA}
 An NFA over $E$ is a tuple $\mathcal{A} = (Q, E, \delta, q_0, F)$ with a finite set of states $Q$; alphabet $E$; transition function $\delta: Q \times E \rightarrow \pow{Q}$; 
initial state $q_0$; and a set of accepting (final) states $F \subseteq Q$.
\end{definition}
%
A state sequence $q_0 q_1 \cdots q_n \in Q^*$ where $q_0$ is the initial state is, a \emph{trace} for a word  $r = e_1 e_2 \cdots e_n \in E^*$ if $q_{i+1} \in \delta(q_i, e_i)$ for each $i \in \{0, \ldots, n-1 \}$.
The NFA \emph{accepts} word $r=e_1 \cdots e_n$ iff there is a trace $q_0 \cdots q_n$ for $r$ s.t. $q_n \in F$.
The finite words accepted by the NFA constitute the language of the NFA, denoted $L(\mathcal{A})$, i.e., $L(\mathcal{A})=\{ r \in E^* \mid r \ \text{is accepted by } \mathcal{A} \}$.
%
%
A deterministic finite automaton (DFA) is an NFA $\mathcal{A} = (Q, E, \delta, q_0, F)$ in which for each $q \in Q$ and $a \in E$, $|\delta(q, a)| = 1$.
Accordingly, the transition function of a DFA is treated as $\delta: Q \times E \rightarrow Q$.
A DFA whose transition function is partial is called a partial DFA.
%


\subsection{Problem Description}
The system allows the agent to take only specific tours in the environment. 
\begin{definition}[Itinerary DFA]
    The itinerary DFA for $\mathcal{G} = (V, E, \src, \tgt, v_0, S,\mathbb{Y},\mathcal{O})$ is a DFA $\mathcal{I} = (Q_\mathcal{I}, E, \delta_{\mathcal{I}}, q_0^\mathcal{I}, F_{\mathcal{I}})$. 
\end{definition}
%

%
Each word accepted by this DFA, if it is a valid walk over the world graph, represents a single tour the agent is allowed to take.
This word specifies not only the locations along the tour but also the specific transitions that are taken along the tour.
%

When the agent moves in the environment, the system does not know the exact tour taken, but it receives world-observations instead,
which uses to verify if the agent followed the itinerary.
%
During a transition $e$, all the events in $\mathcal{O}(e)$ occur simultaneously. 
%
%
The agent's tour, represented by a walk over $\mathcal{G}$, generates a sequence of non-empty world-observations.
Abusing the notation, we use $\mathcal{O}: \walks(\mathcal{G})  \rightarrow (\pow{Y} \setminus \emptyset)^*$ to map a walk $r  =  e_1 \cdots e_n \in \walks(\mathcal{G})$ to its corresponding sequence of world-observations $\mathcal{O}(r) = z_1 \cdots z_{n}$ in which $z_i  = \mathcal{O}(e_i)$ for each $i \in \{1, \ldots, n \}$.
We define conditions enabling the system to verify if the agent followed the itinerary.
\begin{definition}[Certifying Sensor Set]
    \label{def:legSens}
    Sensor set $S$ \emph{certifies} itinerary $\mathcal{I}$ on world graph $\mathcal{G}$ if there exist no $r \in L(\mathcal{I}) \cap \walks(\mathcal{G})$ and $t \in \walks(\mathcal{G}) \setminus L(\mathcal{I})$ s.t. $\mathcal{O}(r) = \mathcal{O}(t)$.
\end{definition}
%

In words, $S$ is certifying if for any walk the agent takes, the system can verify, solely based on the world-observation sequence it receives, whether the agent's walk was within the itinerary.
%
%
The polynomial-time algorithm in \cite{rahmani2021sensor} does that verification.
%
%
%
\vspace{-6pt}
\begin{assumption}
Sensor set $S$ within the given world graph $\mathcal{G}$ is certifying for $\mathcal{I}$.
\end{assumption}
%

%
We consider an attacker, or an adversarial agent, who intends to deviate from the allowed itinerary. 
%
Its deviation itinerary is specified by a DFA.
%
\begin{definition}[Deviation DFA]
    A \emph{deviation DFA} for $\mathcal{G} = (V, E, \src, \tgt, v_0, S,\mathbb{Y},\mathcal{O})$ is a DFA $\mathcal{D} = (Q_{\mathcal{D}}, E, \delta_{\mathcal{D}}, q_{0}^{\mathcal{D}}, F_\mathcal{D})$.
\end{definition}

The agent's objective is to be able to take any tour within $L(\mathcal{D})$ without being detected by the system. Because in general $L(\mathcal{D}) \setminus L(\mathcal{I}) \ne \emptyset$, the agent employs sensor alteration attacks to mislead the system into believing the tour it took is allowed by $\mathcal{I}$.

We make the following assumption about the adversarial agent's sensor alteration:
\vspace{-6pt}
\begin{assumption}
    The sensor alteration occurs before taking any tour.
\end{assumption}
In words, the agent cannot dynamically decide what sensor readings to alter. 
%
%

%
 \begin{definition}[Sensor alteration]
\label{def:cost_func}
A sensor alteration is a function $A: Y \rightarrow Y $ where for each
$y \in Y$, $A(y)$ is the sensor reading triggered by the sensors under attack whenever the sensors would trigger event $y$ when they were not under attack. 
  The sensor alteration cost function is a function $c: Y \times Y  \rightarrow \mathbb{R}_{\geq 0} \cup \{ \infty \}$ where for each $y_1, y_2 \in Y$, $c(y_1, y_2)$ is the cost for the agent to alter event $y_1$ to event $y_2$.
\end{definition}
%
%
\vspace{-6pt}
Case $c(y_1, y_2) = \infty$ means the agent cannot convert event $y_1$ into event $y_2$.

Let $\mathbf{A}$ be the set of all sensor alterations. Function $C: \mathbf{A} \rightarrow \mathbb{R}_{\geq 0} \cup \{ \infty \}$ gives for each sensor alteration $A \in \mathbf{A}$, the cost associated with $A$ as $C(A) = \sum_{y \in Y} c(y, A(y))$.
%

We use $\mathcal{O}_A: E \rightarrow \pow{Y}$ to indicate for each $e \in E$, the world-observation produced by the sensors under the sensor alteration $A$ for $e$, i.e., $\mathcal{O}_A(e) = \bigcup_{y \in \mathcal{O}(e)} \{A(y)\}$.
Abusing the notation, we also let $\mathcal{O}_A: \walks({\mathcal{G}}) \rightarrow (\pow{Y} \setminus \emptyset)^*$ be a function such that for each
 walk $r=e_1 e_2 \cdots e_n \in \walks(\mathcal{G})$,  $\mathcal{O}_A(r)$ is the sequence of world-observations the system receives under the sensor alteration $A$, i.e., $\mathcal{O}_A(r) = z_1 z_2 \cdots z_n$ in which for each $i \in \{1, 2, \cdots, n \}$,
$z_i = \mathcal{O}_A(e_i)$. 
%

%
The goal is to convince the system that the agent did not deviate from the itinerary.
%
\begin{definition}
\label{def:decept_sens}
A sensor alteration attack $A: Y \rightarrow Y$ is \emph{deceptive} if for each walk $r \in L(\mathcal{D}) \cap \walks(\mathcal{G})$, there exists
a walk $r' \in L(\mathcal{I}) \cap \walks(G)$ such that $\mathcal{O}_A(r) = \mathcal{O}(r')$.
\end{definition}
%
%
Intuitively, with a deceptive sensor alteration, any walk within the deviation produces a world-observation sequence that can be produced by at least one walk
within the itinerary.
Therefore, if the agent does a deceptive sensor alteration attack, then it can take any of its intended walks without being detected by the system.
%
%
        \problemdef{Minimum-cost sensor deception (\MCSD)}
{A world graph $\mathcal{G} = (V, E, \src, \tgt, v_0, S,\mathbb{Y},\mathcal{O})$, an itinerary DFA $\mathcal{I} = (Q_{\mathcal{I}}, V, \delta, q_{0, \mathcal{I}}, F_{\mathcal{I}})$ for which $S$ is certifying, a deviation DFA $\mathcal{D} = (Q_{\mathcal{D}}, V, \delta_{\mathcal{D}}, q_{0, \mathcal{D}}, F_{\mathcal{D}})$, and a sensor alteration cost function $c: Y \times Y \rightarrow \mathbb{R}_{\geq 0} \cup \{ \infty \}$.
}
{A deceptive sensor alteration \mbox{$A$} that minimizes cost function $C$, or `\textsc{Infeasible}' if no deceptive sensor alteration exists.}
    \vspace{-\baselineskip}
        \section{Hardness of \MCSD}
        \label{sec:hard}
         We present a hardness result, for which
         we first define our problem's decision variant.
          %

                \decproblem{Minimum-cost sensor deception (\MCSDDEC)}
{World graph $\mathcal{G}$, itinerary DFA $\mathcal{I}$, deviation DFA $\mathcal{D}$, and sensor alteration cost function $c$, defined in \MCSD, and a non-negative real number $l \in \mathbb{R}_{\geq 0}$
}
{\Yes if there is a deceptive sensor alteration \mbox{$A$} s.t. $C(A) \leq l$, and \No otherwise.}
         
         Then we review  a well-known NP-hard problem. 
                         \decproblem{Directed Multi-cut (\DMCDEC)}
{A directed graph $G = (V', E')$, an indexed set of source-target pairs $T=\{(s_1, t_1), (s_2, t_2), \cdots, (s_n, t_n) \} \subseteq V' \times V'$, and a non-negative integer $k$.
}
{$\Yes$ if there exists a $T$-cut---a set of edges $O \subseteq E'$ whose removal from $G$ disconnects $t_i$ from $s_i$ for all $(s_i, t_i) \in T$)--- such that $|O| \leq k$, and \No otherwise.}
    This problem is known to be $\NP$-hard.
    We need a special variant of this problem.

    \decproblem{Connected Directed Multi-cut (\CDMCDEC)}
{A directed graph $G = (V', E')$, an indexed set of source-target pairs $T=\{(s_1, t_1), (s_2, t_2), \cdots, (s_n, t_n) \} \subseteq V' \times V'$, where for each $i \in \{1, 2, \cdots, n\}$, there exists at least a path from $s_i$ to $t_i$, and a non-negative integer $k$.
}
{$\Yes$ if there exists a $T$-cut $O$ such that $|O| \leq k$, and \No otherwise.}
The difference between the general multi-cut problem and this special variant is that in the later each source state $s_i$ is connected to the target state $s_t$.
%
%
\begin{lemma}
    \label{lem:conMultiCut_NPHard}
          \CDMCDEC $\in \NP$-hard.
\end{lemma}
\vspace{-14pt}
\begin{proof}
    By reduction from the general multi-cut problem.
    The idea is to add an edge $(s, t)$ for each source-target $(s, t) \in T$ for which 
    the graph has no path from $s$ to $t$.
    Formally, given an \DMCDEC instance 
    \begin{equation}
	      \langle G := (V', E'), T=\{(s_1, t_1), (s_2, t_2), \cdots, (s_n, t_n) \}, k \rangle,    
	    \end{equation}
     we make a \CDMCDEC instance 
     \begin{multline}
         \langle G_2 := (V', E' \cup E''), T=\{(s_1, t_1), (s_2, t_2), \cdots, (s_n, t_n) \}, k+|E''| \rangle, 
     \end{multline}
     where $E'' = \{ (s, t) \in T \mid \text{there is no path from $s$ to $t$ in $G$}\}$.
     Clearly, this reduction takes a polynomial time. 
     It is also easy to prove that the reduction is correct, that is, \DMCDEC has a $T$-cut of size at most $k$ if and only if \CDMCDEC has a $T$-cut of size at most $k+|E''|$.
     The idea is that the union of any $T$-cut for \DMCDEC with $E''$ is a $T$-cut for the \CDMCDEC instance, and that
     any $T$-cut $O$ for the \CDMCDEC instance contains $E''$, and thus, $O \setminus E''$ is a $T$-cut for the  \DMCDEC instance.
     %

     
     %
\end{proof}
    
\begin{figure*}[t!]
  \centering
  \includegraphics[width=1.0\linewidth]{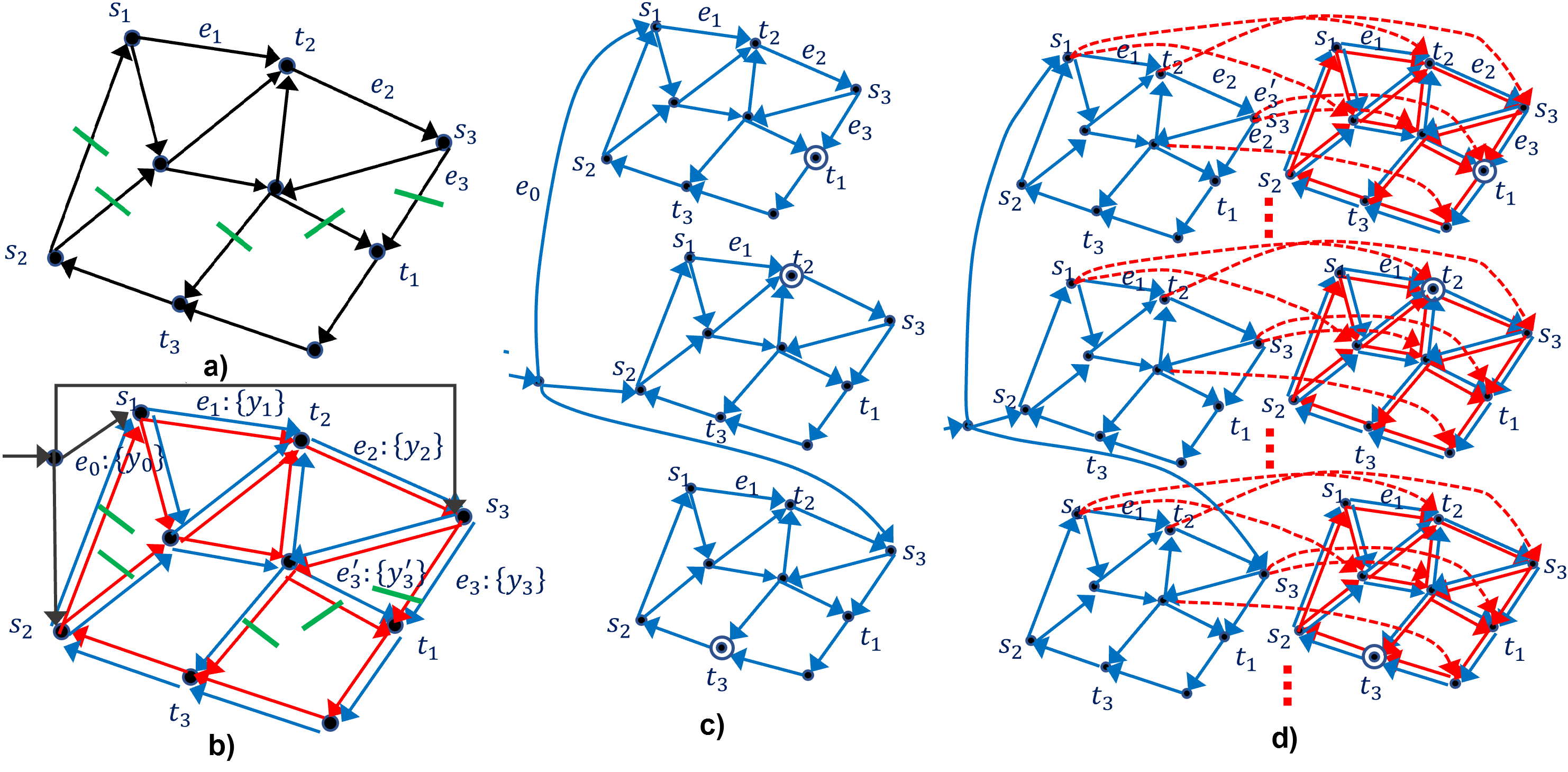}
  \vspace{-20pt}
  \caption{
    Reduction from the multi-cut problem to our problem. 
    \textbf{a)} An instance of the multi-cut problem, which has three $(s_i, t_i)$' source-target pairs. The edges in the minimal cut are crossed with green segments.
    \textbf{b)} The world-graph of the instance of our problem constructed by the reduction. Edge names are omitted to improve readability.
                The sensor events of only those red edges crossed by the green segments are altered.
                Each of those events converted into the event produced by the twin blue edge of the red edge producing that event.
    \textbf{c)} The itinerary DFA $\mathcal{I}$ constructed by the reduction. All the mission edges, including the red edges, enter a trapping state omitted to reduce visual clutter.
                Each row has only a single accepting state, $t_1$, $t_2$, and $t_3$ in those three rows, respectively.
    \textbf{d)} The DFA $\mathcal{D}$, constructed by the reduction. Each dotted edge connects a vertex from the left column to the right column and each corresponds to a red edge in the world-graph (left column). 
    Some of those red edges from the left column to the right column are omitted to reduce visual clutter.
    The trapping state $q_{trap}$ and all the transitions entering that state are omitted too.
    The DFA has only three accepting states, drawn by double circles.
  \label{fig:reduction}}
  \vspace{-14pt}
\end{figure*}

    We use this to prove that our sensor deception problem is computationally hard.
    \begin{theorem}
        \label{thr:NPHard}
          \MCSDDEC $\in \NP$-hard.
    \end{theorem}
    \vspace{-14pt}
		\begin{proof}
		By reduction from the \CDMCDEC problem.
        The idea is to construct for graph $G$ within in an instance of the \CDMCDEC, a world-graph $\mathcal{G}$ for the instance of the \DMCDEC where $\mathcal{G}$ has the same vertex set of $G$ with an additional vertex as the initial vertex. Each edge $e$ in $G$ is duplicated as two parallel edges, one blue and one red, considered as ``twins'' in $\mathcal{G}$. For each edge created for $\mathcal{G}$, one distinct sensor triggering a distinct event is assigned.
        Altering an event assigned to a red edge to its twin's event incur a cost of 1, while any other alterations incurs an infinite cost.
        See Fig. ~\ref{fig:reduction}.
        The allowed itinerary consists of the paths connecting the $s_i$'s to the $t_i$'s via only the blue edges.
        The deviation would consist of all the paths that connect the $s_i$'s to the $t_i$'s such that each path must pass through at least one red edge.
        From a sensor alteration for the \DMCDEC problem, a set of red edges are chosen and those edges in $G$ for which those red edges were created constitute as a $T$-cut for the \CDMCDEC instance.
        %

        
	    Formally, given a \CDMCDEC instance 
	    \begin{equation}
	      \langle G := (V', E'), T=\{(s_1, t_1), (s_2, t_2), \cdots, (s_n, t_n) \}, k \rangle,    
	    \end{equation}
	    we make a \MCSDDEC instance
	    \begin{align*}
	        \langle \mathcal{G} := (V, E, \src, \tgt, v_0, S,\mathbb{Y},\mathcal{O}), 
	        \mathcal{I} := (Q_{\mathcal{I}}, E, \delta_{\mathcal{I}}, q_{0, \mathcal{I}}, F_{\mathcal{I}}), \\
	        \mathcal{D} := (Q_{\mathcal{D}}, E, \delta_{\mathcal{D}}, q_{0, \mathcal{D}}, F_{\mathcal{D}}),
	        c,
	        l
	        \rangle
	    \end{align*}
	    in which (a) we construct $\mathcal{G}$ as follows
	    \begin{itemize}
	        \item we make a set $\mathbf{B}$ for blue edges and a set $\mathbf{R}$ for red edges and set
	          $\mathbf{B} = \mathbf{R} =  E = \emptyset$, 
	        \item we make an initial state $v_0$ and set $V = V' \cup \{v_0 \}$,
	        \item for each $e = (v_1, v_2) \in E'$, we make two edges $e_1, e_2$ s.t. $\src(e_1)=\src(e_2)=v_1$ and 
	        $\tgt(v_1)=\tgt(v_2) = v_2$, add $e_1$ and $e_2$ to $E$, add $e_1$ to $\mathbf{B}$, add $e_2$ to $\mathbf{R}$, and 
	        we call $e_1$ and $e_2$ twins of each other, i.e., $\twin(e_1) = e_2$ and $\twin(e_2)=e_1$,
	        \item for each $(s_i, t_i) \in T$, we add an edge $e$ to $E$ with $\src(e)=v_0$ and $\tgt(e)=s_i$,
	        \item for each edge $e \in E$, we make a sensor $s_e$ and add $s_e$ to $S$ and we use $\edge(s_e)$ to denote $e$---the edge $s_e$ was created for, and
	        \item for each sensor $s_i \in S$, assuming $e = \edge(s_i)$ to be the edge $s_i$ was made for (see above), we make an observation $y_i$ and set $\mathcal{O}(e) = \{ y_i \}$, and because the observation function for this reduction is a one-to-one function, it holds that $\mathcal{O}^{-1}(y_i) = e$;
	    \end{itemize}
	    (b) we construct $\mathcal{I} := (Q_{\mathcal{I}}, E, \delta_{\mathcal{I}}, q_{0, \mathcal{I}}, F_{\mathcal{I}})$ as follows
	    \begin{itemize}
	        \item $Q_{\mathcal{I}} = \{(v_0, 0)\} \cup \{ (v, i) \mid v \in V, i \in \{1, \cdots, |T| \} \} \cup \{q_{trap} \}$,
	        \item $q_{0, \mathcal{I}} = (v_0, 0)$,
	        \item for each $q=(v, i) \in Q_{\mathcal{I}}$ and $e \in E$, 
        \begin{equation}
            \delta_{\mathcal{I}}((v, i), e) =
			\begin{cases}
			(\tgt(e), i) & \src(e)=v \text{ and } e \in \textbf{B} \\
			
			(\tgt(e), i+1) & \src(e)=v \text{ and } v=v_0  \\
			
			q_{trap} & \text{otherwise},
			
			\end{cases}    
            \end{equation}
            \item for each $e \in E$, $\delta_{\mathcal{I}}(q_{trap}, e) = q_{trap}$,
	        \item $F = \{ (t_i, i) \mid  i \in \{1, \cdots, |T| \} \}$,
         
         with the intuition that $\mathcal{I}$ makes $|T|$ copies of $G$ using only the blue edges where in each copy exactly one target state $t_i$ is made to be accepting, and all other edges, including all the red edges, are transitioned to the trapping state,
	    \end{itemize}
	    (c) we construct $\mathcal{D} = (Q_{\mathcal{D}}, E, \delta_{\mathcal{D}}, q_{0, \mathcal{D}}, F_{\mathcal{D}})$ as follows
	    \begin{itemize}
	        \item $Q_{\mathcal{\mathcal{D}}} = \{(v_0, 0, 0)\} \cup \{ (v, i, j) \mid v \in V, i \in \{1, \cdots, |T| \}, j \in \{1, 2 \} \}
	         \cup \{q_{trap}\}
	        $
	        \item $q_{0, \mathcal{D}}=(v_0, 0, 0)$,
	        \item for each $q=(v, i, j) \in Q_{\mathcal{D}}$ and $e \in E$, 
        \begin{equation}
            \delta_{\mathcal{D}}((v, i, j), e) =
			\begin{cases}
			(\tgt(e), i, j) & \src(e)=v \text{ and }  (e \in \textbf{B} \text{ or } j=2) \\
			
			(\tgt(e), i, j+1) & \src(e)=v \text{ and }  e \in \textbf{R} \text{ and } j=1 \\
			
			(\tgt(e), i, 1) & \src(e)=v \text{ and } v=v_0  \\
			
			q_{trap} & \text{otherwise},
			
			\end{cases}    
            \end{equation}
            \item for each $e \in E$, $\delta_{\mathcal{D}}(q_{trap}, e) = q_{trap}$,
            \item $F_{\mathcal{D}} = \{ (t_i, i, 2) \mid i \in \{1, \cdots, |T| \}\}$,
            with the intuition that $\mathcal{D}$ copies $G$ into $|T|$ rows and two columns where the first column is used only for the blue edges while the second column contain both the blue edges and the red edges and the second column is reached from the first column by red edge and that at each row there is exactly one accepting state $t_i$, located in the second column, 
            \end{itemize}
	        and (d) we define $c$ such that for each $e_1 \in E$ and $e_2 \in E \cup \{\epsilon\}$, $c(e_1, e_2) = 1$ if $e_1 \in \mathbf{R}$ and $e_2 = \twin(e)$, and otherwise $c(e_1, e_2) = \infty$; 
	        and finally, (e) we set $l = k$.
	    
	       Fig.~\ref{fig:reduction} illustrates this reduction.
	    The language of $\mathcal{I}$ consists of paths starting from $v_0$, connecting $s_i$'s to their corresponding $t_i$'s via only blue edges.
        The language of $\mathcal{D}$ consists of all such paths where each path passes through at least a red edge.
        Essentially, the system permits the agent to to solely use blue edges, requiring it to end at $t_i$ if it has entered $s_i$, starting at $v_0$.
        The sensor attack aims to map the events for a set of red edges into the events of their corresponding twin edges. 
         As these twin edges are all blue, the system remains unaware of the deviation since no event for a red edge is triggered in its perception and
      the world-observation sequence produced by a deviated path after sensor attack matches the world-observation sequence of a path in the itinerary. 
        
        Clearly, the reduction takes a polynomial time.
        We need only to show the correctness of the reduction. We prove that \MCSDDEC has a solution of size at most $l=k$ if and only if \CDMCDEC has a solution of size at most size $k=l$.

        ($\Rightarrow$) Suppose there exists a set of edges $O \subseteq E'$ whose removal from $E'$ disconnects $t_i$ from $s_i$ for each integer $1 \leq i \leq |T|$ and that $|O|=k' \leq k$.
        Notice that for each edge $e \in E'$, we created and added two parallel edges $e_1$ and $e_2$, one blue and the other red, to the world-graph $\mathcal{G}$.
        Now for each edge $e \in O$, let $\red(e)$ and $\blue(e)$ be respectively its corresponds blue and red edges in $\mathcal{G}$.
        In the sensor attack, we map the sensor event created for the red edge $\red(e)$ into the sensor event for the blue edge $\blue(e)$.
        Formally, for each edge $e \in E'$, we let $A(\mathcal{O}^{-1}({\red(e)})) = \mathcal{O}^{-1}({\blue(e)})$ and $A(\mathcal{O}^{-1}({\blue(e)})) = \mathcal{O}^{-1}({\blue(e)})$ if $e \in O$, and otherwise, $A(\mathcal{O}^{-1}({\red(e)})) = \mathcal{O}^{-1}({\red(e)})$ and $A(\mathcal{O}^{-1}({\blue(e)})) = \mathcal{O}^{-1}({\blue(e)})$.
        Clearly, $A$ is a deceptive sensor attack for $\mathcal{G}$, $\mathcal{I}$, and $\mathcal{D}$ and that $C(A) = k' \leq k=l$.
        
        ($\Rightarrow$) Conversely, assume there exists a sensor attack $A$ for the instance of \MCSDDEC such that $C(A) = l' <= l$.
        By the construction of the cost function $c$, the only way to alter the sensor events with a finite cost is when we map the event produced by a red edge
        into the event produced by a blue edge.
        Formally, let $y_1, y_2 \in Y$ be two events such that $0 < c(y_1, y_2) \neq \infty$.
        Trivially, $e_1 = \mathcal{O}^{-1}(y_1) \in \mathbf{R}$, that is, $e_1$ is a red edge.
        For each such red edge $e_1$, let $\gray(e_1)$ be the edge in graph $G$ for which the red edge $e_1$ was created.
        We let $O = \{\gray(\mathcal{O}^{-1}(y_1)) \mid y_1 \in Y \text{ and } c(y_1, A(y_1))=1 \}$.
        Clearly, $O$ is a multi-cut for the \DMCDEC problem and that $|O|=l' \leq l=k$.
        %
		\end{proof}

   %
   To see the connection between a $T$-cut in \CDMCDEC and a sensor alteration in \DMCDEC, consider the simple path
   $e_1 e_2 e_3$ in the graph in Fig.~\ref{fig:reduction}a, which connects $s_1$ to $t_1$.
   This path is cut by the chosen $T$-cut, shown as the edges crossed by the line segments in green,
   because edge $e_3$ is contained in the $T$-cut.
    The walk corresponding to this path is the walk $e_1 e_2 e_3$ shown in Fig.~\ref{fig:reduction}b.
    This walk is in the allowed itinerary.
    Walk $e_1 e_2 e_3'$, shown in the same figure, is within the deviation itinerary.
    The sensor alteration, which is shown by the green segments crossing the red edges whose corresponding events are altered, opts to convert $y_3'$ into $y_3$.
    Therefore, when the agent takes this walk and the system is under attack, the sequence of world-observation produced for this walk is ${y_1} {y_2} {y_3}$, which is produced for walk $e_1 e_2 e_3$ when the system is not under attack.
   %

%
        %
        %
         %
        %
        \vspace{-0.5\baselineskip}
        \section{Sensor Alteration Verification}
        \label{sec:verif}
        In this section, we provide an algorithm for verifying whether a given sensor alteration is deceptive in the sense of Definition~\ref{def:decept_sens}.
        We first define a product automata construction. 
         %
        %
         %
         \vspace{-0.4\baselineskip}
         \begin{definition}[Product automaton]
         \label{def:prodAut}
           Given a world-graph $\mathcal{G} = (V, E, \src, \tgt, v_0, S,\\ \mathbb{Y},\mathcal{O})$ and a DFA $\mathcal{A} = (Q, E, \delta, q_0^\mathcal{A}, F)$, \emph{their product} $\mathcal{P} = \mathcal{A} \times \mathcal{G}$ is 
           a partial DFA $\mathcal{P} = (Q_{\mathcal{P}}, E, \delta_{\mathcal{P}}, q_0^{\mathcal{P}}, F_{\mathcal{P}})$ where 
        \begin{inparaenum}[(1)]
               \item $Q_{\mathcal{P}} = Q \times V$ is the state space,       
               \item $\delta_{\mathcal{P}}: Q_{\mathcal{P}} \times E \nrightarrow Q_{\mathcal{P}}$, which is a partial function, is the transition function such that for each $(q, v) \in Q_{\mathcal{P}}$ and $e \in E$, if 
               $\srcfunc{e} \neq v$, set $\delta_{\mathcal{P}}((q, v), e) = \bot$, otherwise, $ \delta_{\mathcal{P}}((q, v), e) = (\delta(q, e), \tgtfunc{e})$,
               \item $q_0^{\mathcal{P}}=(q_0, v_0)$, and
               \item $F_{\mathcal{P}}=F \times V$ is the set of final (accepting) states.
           \end{inparaenum}
         \end{definition}
         \vspace{-0.4\baselineskip}
          For a partial DFA, $\delta_{\mathcal{P}}(p, e)=\bot$ means the transition function is undefined for $p \in Q_{\mathcal{P}}$ and $e \in E$.
          The extended transition function $\delta_{\mathcal{P}}^*: Q_{\mathcal{P}} \times E^* \nrightarrow Q_{\mathcal{P}}$ is defined as usual.

        We use two product automata to define the condition that must be satisfied for a sensor alteration to be deceptive.
         \begin{lemma}
         \label{lem:prodAut}
         Let $\mathcal{P}= \mathcal{I} \times \mathcal{G} = (Q_{\mathcal{P}}, E, \delta_{\mathcal{P}}, q_0^{\mathcal{P}}, F_{\mathcal{P}})$ and $\mathcal{M} = \mathcal{D} \times \mathcal{G} = (Q_{\mathcal{M}}, E, \delta_{\mathcal{M}}, \\q_0^{\mathcal{M}}, F_{\mathcal{M}})$.
         A sensor alteration $A: Y \rightarrow Y$ is deceptive for $\mathcal{I}$ and $\mathcal{D}$ if
         for each walk $r \in L(\mathcal{M})$, there is a walk $r' \in L(\mathcal{P})$ such that
         $\mathcal{O}_A(r) = \mathcal{O}(r')$.
         \end{lemma}
         \vspace{-8pt}
         \begin{proof}
           Combine Definition~\ref{def:decept_sens}, that $L(\mathcal{P}) = L(\mathcal{I}) \times \walks(\mathcal{G})$ and that $L(\mathcal{M}) = L(\mathcal{D}) \times \walks(\mathcal{G})$.
          \end{proof}
        As a result, the problem of verifying whether a sensor alteration is deceptive is reduced to the problem of language inclusion for two NFAs 
        we define in the following.

        %
        First, we define the alphabet of those NFAs.
         It must contain all the possible world-observations that can be perceived by the system for both when the system is under a sensor attack
         and when the system is not under a sensor attack.
         %
        %
        Each letter of this alphabet is a multi-set of events rather than a set of events, although in Definition~\ref{def:wg}, we assumed the observation assigned to each edge is a set of events.
        This is due to potential mapping of multiple events to a single event during sensor alterations. 
        %
        For example, assume $\mathcal{O}(e) = \{y_1, y_2, y_3\}$ and $A = \{(y_1, y_1), (y_2, y_1), (y_3, y_3)\}$. In this case, $\mathcal{O}_A(e) = \{y_1, y_1, y_3 \}$, which is a multi-set.        
            Let $m$ be the maximum number of events that can happen simultaneously in the environment, i.e.,
        $m = \max_{e \in E} |\mathcal{O}(e)|$.
        We define the alphabet $\Sigma$ as the set of event multi-sets of size at most $m$, excluding $\emptyset$.
        In the example in Fig.~\ref{fig:department}, $m=2$. 
        Given this $m$, for $Y = \{y_1, y_2, y_3\}$, $\Sigma= \{\{y_1\}, \{y_2\}, \{y_3\},  \{y_1, y_1\}, \{y_2, y_2\}, \{y_3, y_3\},  \{y_1, y_2\}, \{y_1, y_3\}, \{y_2, y_3\} \}$.
        %
        %
        \begin{definition}
        \label{def:nfas}
        %
        The \emph{observation-relaxation} of the DFA  $\mathcal{P}=(Q_{\mathcal{P}}, E, \delta_{\mathcal{P}}, q_0^{\mathcal{P}}, F_{\mathcal{P}})$, defined in Lemma~\ref{lem:prodAut}, and the observation-relaxation of the DFA $\mathcal{M}=(Q_{\mathcal{M}}, E, \delta_{\mathcal{M}}, q_0^{\mathcal{M}}, \\ F_{\mathcal{M}})$ under sensor alteration $A$, are respectively 
        NFAs $\mathrm{P}=(Q_{\mathrm{P}}, \Sigma, \delta_{\mathrm{P}}, q_0^{\mathrm{P}}, F_{\mathrm{P}})$ and 
        $\mathrm{M}=(Q_{\mathrm{M}}, \Sigma, \delta_{\mathrm{M}}, q_0^{\mathrm{M}}, F_{\mathrm{M}})$ 
        in which for $q \in Q_{\mathrm{P}}$ and $x \in \Sigma$,
        \begin{equation}
            \delta_{\mathrm{P}}(q, x) =\bigcup_{e \in E: \delta_{\mathcal{P}}(q, e) \neq \bot,  \mathcal{O}(e) = x} \{ \delta_{\mathcal{P}}(q, e) \},
        \end{equation}
        and for each $q \in Q_{\mathrm{M}}$ and $x \in \Sigma$,
        \begin{equation}
            \delta_{\mathrm{M}}(q, x) =\bigcup_{e \in E: \delta_{\mathcal{M}}(q, e) \neq \bot,  \mathcal{O}_A(e) = x} \{ \delta_{\mathcal{M}}(q, e) \}.
        \end{equation}
        \end{definition}

         %
         \vspace{-8pt}
         While $\mathcal{P}$ and $\mathcal{M}$ are acceptors for sequences of edges in the world-graph, $\mathrm{P}$ and $\mathrm{M}$ are acceptors for
         sequences of world-observations.
         The language of $\mathrm{P}$ consists of all the sequences of world-observations produced by the walks accepted by the itinerary DFA 
         and the language of $\mathrm{M}$ consists of all the sequences of world-observations produced by the walks accepted by the deviation DFA under sensor alteration $A$.
        Thus, the problem of checking if $A$ is deceptive is reduced to the NFA inclusion problem of whether $L(\mathrm{M}) \subseteq L(\mathrm{P})$.
        The NFA inclusion problem is in general PSPACE-Complete~\cite{meyer1972equivalence}. 
        %
        
        Given these NFAs, we present a property of a deceptive sensor alteration.
        %
        \begin{corollary}
        \label{cor:decept_sensor}
        Sensor alteration $A$ is deceptive iff $\overline{L(\mathrm{P})} \cap L(\mathrm{M}) = \emptyset$.
        \end{corollary}
        \vspace{-8pt}
        \begin{proof}
         By Lemma~\ref{lem:prodAut}, $A$ is deceptive if $L(\mathrm{M}) \subseteq L(\mathrm{P})$.
         Also, it is easy to observe given two sets $X$ and $Y$, $X \subseteq Y$ iff $\overline{Y} \cap X = \emptyset$.
         %
         Thus, if $\overline{L(\mathrm{P})} \cap L(\mathrm{M}) = \emptyset$, then $A$ is deceptive.
         Also, if $A$ deceptive, then by Definition~\ref{def:decept_sens}, it means for every walk $r$ within the deviation, there exists a walk $r'$ within the itinerary such that $r$ and $r'$ produce the same sequence of world-observations. 
         Because $\mathrm{P}$ is a DFA with a total transition function, tracing the sequence of world-observations produced for $r$ in $\mathrm{P}$ reaches exactly one state and that state is accepting. This means $L(\mathrm{M}) \subseteq L(\mathrm{P})$, and thus, $\overline{L(\mathrm{P})} \cap L(\mathrm{M}) = \emptyset$.
        \end{proof}
        %
        
        %
            \vspace{-20pt}
          \section{Planning sensor alteration attacks}
          \label{sec:ILP}
        In this section, we present our algorithm to solve \MCSD.
        %
        The algorithm first constructs a DFA accepting all world-observation sequences produced by the walks outside the itinerary.
        For each walk $r$ accepted by this DFA, either $r \in \walks(\mathcal{G}) \setminus L(\mathcal{I})$ or 
        $r \in E^* \setminus \walks(\mathcal{G})$, that is, either $r$ is realizable by the world-graph but is not within the itinerary or it is not realizable by the world-graph.
        This DFA is constructed by converting the NFA $\mathrm{P}=(Q_{\mathrm{P}}, \Sigma, \delta_{\mathrm{P}}, q_0^{\mathrm{P}}, F_{\mathrm{P}})$, introduced in Definition~\ref{def:nfas}, into a DFA using the standard determinization method~\cite{hopcroft2001introduction}, 
        and then converting the accepting states of this DFA into non-accepting states, and vice versa.
        To this end, we obtain a DFA $\mathrm{O}=(Q_{\mathrm{O}}, \Sigma, \delta_{\mathrm{O}}, q_0^{\mathrm{O}}, F_{\mathrm{O}})$ accepting the language $\overline{L(\mathrm{P})}$, that is, 
    $L(\mathrm{O}) = \overline{L(\mathrm{P})}$.
        The algorithm then uses $\mathrm{O}$ along with the deviation DFA $\mathcal{M}=(Q_{\mathcal{M}}, E, \delta_{\mathcal{M}},  q_0^{\mathcal{M}}, F_{\mathcal{M}})$ in an 
        integer linear programming (ILP) problem that minimizes the cost of sensor alteration while making sure that the intersection of $L(\mathrm{O})$ and $L(\mathrm{M})$ is the empty set (see Corollary~\ref{cor:decept_sensor}).  
        The ILP uses $\mathcal{M}$ to simulate $L(\mathrm{M})$ on the fly.
        %
        %
        %
        %
        %
        The following result reveals the main idea of this integer linear program.
        %
        
\begin{figure*}[t]
  \centering
  \includegraphics[width=0.6\linewidth]{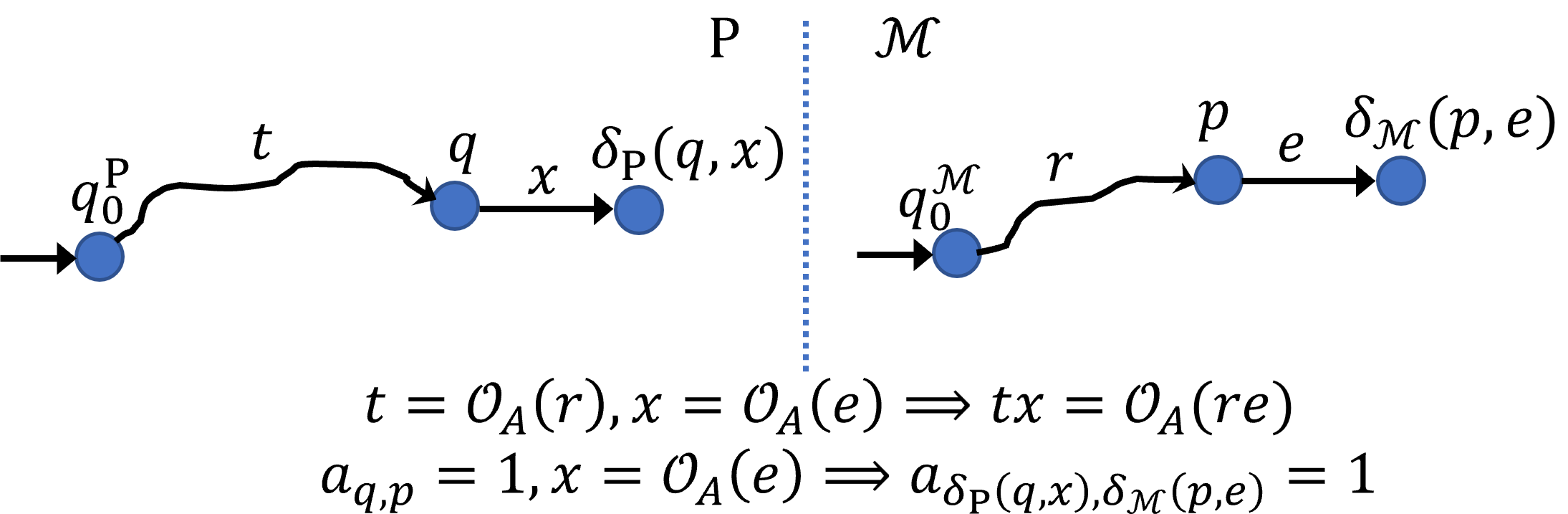}
  \vspace{-10pt}
  \caption{
    Main idea of our ILP. Given that walk $r$ produces, under the sensor attack, the world-observation sequence $t$, and that $r$ reaches $p$ in $\mathcal{M}$ and $t$ reaches $q$ in $\mathrm{O}$, $a_{q, p} = 1$. Edge $e$ yields, under the sensor alteration, the world-observation $x$. Given these, $a_{\delta_{\mathrm{O}}(q, x), \delta_{\mathcal{P}}(p, e)} = 1$.
      \label{fig:ilp_rule}}
\vspace*{-18pt}
\end{figure*}

        %
        \begin{corollary}
        \label{cor:ilp_idea}
        Sensor alteration $A$ is deceptive iff there exist no walk $r \in \walks(\mathcal{G})$ and world-observation sequence $t \in \Sigma^*$ s.t. $t = \mathcal{O}_A(r)$, $\delta^*_{\mathcal{M}}(q_0^{\mathcal{M}}, r) \in F_{\mathcal{M}}$, and $\delta^*_{\mathrm{O}}(q_0^{\mathrm{O}}, t) \in F_{\mathrm{O}}$.       
        \end{corollary}
        \vspace{-14pt}
        \begin{proof}
           Combine Lemma~\ref{lem:prodAut}, Corollary~\ref{cor:decept_sensor}, and that $L(\mathrm{O}) = \overline{L(\mathrm{P})}$.
        \end{proof}
        In words, a sensor alteration is deceptive if, under that sensor attack, every walk in the deviation DFA produces a world-observation sequence that could be generated by a walk in the itinerary DFA.
        Accordingly, we introduces a binary variable $a_{q, p}$ for each $q \in Q_{\mathrm{O}}$ and $p \in Q_{\mathcal{M}}$.
        This variable receives value $1$ iff $q$ is reachable by a sequence of world-observations produced by the system under sensor attack for a walk reaching $p$.
        %
        %
        The initial condition is $a_{q_0, p_0}=1$, where $q_0$ is the initial state of $\mathrm{O}$ and $p_0$ is the initial state of $\mathcal{M}$. 
        This is because $q_0$ is reached by the world-observation $\epsilon$, the empty string, which is produced for the walk $\epsilon$ and this walk reaches state $p_0$.
        %
        %
        %
        The following constraint is required: If $a_{q, p}=1$, then for any input $e$ at $p$, it must set $a_{\delta_{\mathrm{O}}(q, x), \delta_{\mathcal{M}}(p, e)} = 1$ where $x$ is the world-observation under the sensor alteration of $e$.
        See Fig.~\ref{fig:ilp_rule}.

        Given Corollary~\ref{cor:ilp_idea}, the sensor alteration is deceptive iff for all $q \in F_{\mathrm{O}}$ and $p \in F_{\mathcal{M}}$, $a_{q, p} = 0$, meaning  any walk in the deviation produces a world-observation sequence that fails to reach the accepting states of $\mathrm{O}$.
        Note the accepting states of $\mathrm{O}$ are reached only by world-observation sequences that are 
        either not realized by the world-graph or not produced for a walk in the itinerary.
        Also, since $\mathrm{O}$ is a DFA with a complete transition function, the world-observation of any walk that can be traced by $\mathcal{M}$ reaches a single state in $\mathrm{O}$.
        If that state is not accepting, then that walk produces, under the sensor alteration, a world-observation sequence producible by a walk within the itinerary. 
        
        %

        The ILP introduces a binary variable $u_{y, y'}$ for each pair of events $y, y' \in Y$, where $u_{y, y'}$ receives $1$ iff the sensor attack alters $y$ into $y'$.
        Also, for each $x \in \Sigma$ and $y \in Y$, an integer variable $n_{y, x}$ is introduced to indicate the multiplicity of $y$ within $x$ (recall that $x$ is a multi-set of events).
            These $n_{y, x}$'s are used to decide if an edge $e \in E$ produces a world-observation $x \in \Sigma$, under the sensor alteration.
            Specifically, $e$ produces $x$ under the sensor alteration iff for each $y \in x$, the number of times the events in $\mathcal{O}(e)$ are mapped to $y$ equals $n_{y, x}$, and no event in $ \mathcal{O}(e)$ is mapped to an event outside $x$.
            For example, if $x = \{ y_1, y_1, y_2\}$ and $\mathcal{O}(e) = \{ y_1, y_2, y_3\}$ with sensor alteration $A= \{ (y_1, y_1), (y_2, y_1), (y_3, y_2)\}$, then $e$ produces $x$ under the sensor alteration.
                However, for $A = \{ (y_1, y_1), (y_2, y_2), (y_3, y_2)\}$, $e$ produces $\{y_1, y_2, y_2\}$, which is not equal to $x$.
            %

        The mathematical program uses these variables as follows.

    \begin{mdframed}
\small
	\noindent Minimize:\vspace{-\baselineskip}
	\begin{equation}  \label{mp:obj}
		\sum_{y \in Y} \sum_{y' \in Y} u_{y, y'} \cdot c(y, y')
	\end{equation}
	Subject to:
	\begin{enumerate}[-]
	     \item For $q = q_0^{\mathrm{O}}$ and $p = q_0^{\mathcal{M}}$, \vspace{-\baselineskip}\begin{equation} \label{mp:init}
			    a_{q, p} = 1  
		    \end{equation}  
		\item $\forall y \in Y$,     \vspace{-\baselineskip}
	        \begin{equation} \label{mp:mapping}
			    \sum_{y' \in Y} u_{y, y'} = 1
		    \end{equation}  	
        \item $\forall y, y' \in Y$ s.t. $c(y, y') = \infty$,  \vspace{-\baselineskip}
	        \begin{equation} \label{mp:infeasible_mapping}
			    u(y, y') = 0
		    \end{equation}  	      
		\item $\forall q \in  F_{\mathrm{O}}$ and $p \in F_{\mathcal{M}}$,       \vspace{-\baselineskip}
	        \begin{equation} \label{mp:deceptive}
			    a_{q, p} = 0
		    \end{equation}  
		\item  $\forall q \in Q_{\mathrm{O}}$, $p \in Q_{\mathcal{M}}$, $x \in \Sigma$, and $e \in E$ s.t. $\delta_{\mathcal{M}}(q, e) \neq \bot$,
		    \begin{align} 
		    \label{mp:edgesSameLabel}
			\notag    a_{q,p} = 1  \wedge (\forall y \in x, \sum_{y' \in \mathcal{O}(e)}u_{y', y}=n_{y, x}) \wedge \\ 
			     (\forall y \in Y \setminus x, \sum_{y' \in \mathcal{O}(e)}u_{y', y}=0)
			    \Rightarrow a_{\delta_{\mathcal{O}}(q, x), \delta_{\mathrm{M}}(p, e)} = 1
		    \end{align} 
		\item $\forall q \in Q_{\mathrm{O}}, p \in Q_{\mathcal{M}}$,     \vspace{-\baselineskip}
	        \begin{equation} \label{mp:a}
			    a_{q,p} \in \{0, 1\}
		    \end{equation}
		\item $\forall y_1 \in Y, y_2 \in Y \cup \{ \epsilon \}$,        \vspace{-\baselineskip}
	        \begin{equation} \label{mp:u}
			    u_{y_1, y_2} \in \{0, 1\}
		    \end{equation}
		  
	\end{enumerate}
\end{mdframed}%

       The objective (\ref{mp:obj}) is to minimize the cost of sensor alteration.
       %
    Constraint (\ref{mp:init}) applies the initial condition of the rule described by Fig.~\ref{fig:ilp_rule}.
    %
    %
    %
   Constraints (\ref{mp:mapping}) indicate that under the sensor alteration, each event $y$ is mapped onto only a single event $y'$. 
    %
    Constraints (\ref{mp:infeasible_mapping}) mean a sensor $y$ cannot be mapped to a sensor $y'$ if the cost of that mapping is infinite. 
    %
    Constraints of type (\ref{mp:deceptive}) guarantee that the chosen sensor alteration is deceptive by applying the result in Corollary~\ref{cor:ilp_idea}.
    %
    Constraints (\ref{mp:edgesSameLabel}) apply the rule described in Fig.~\ref{fig:ilp_rule}.
    %
    %
    %
    %
    Constraints (\ref{mp:a}) and Constraints (\ref{mp:u}) ensure that all the used variables are binary.

    This mathematical program gives the exact solution, but it is still not an integer linear program.
This is because a constraint of type (\ref{mp:edgesSameLabel}) is a logical condition.
To convert the optimization problem into an integer linear program, we employ big-M methods. For each $x \in \Sigma$, $e \in E$, and $y \in x$, we introduce a binary variable $b_{x, e, y}$ which equals to $0$ if and only if $\sum_{y' \in \mathcal{O}(e)}u_{y', y}=n_{y, x}$, that is, if and only if all the occurrences of $y$ within $x$ are mapped to by the events within $\mathcal{O}(e)$.
    %
    
    %
    Accordingly, we introduce the following constraints using the big-M method for a sufficiently large $M$: 
    \begin{mdframed}
\small
	\begin{enumerate}[-]
		\item $\forall x \in \Sigma$, $e \in E$, and $y \in x$, \vspace{-1.8\baselineskip}
	        \begin{align}
	            \sum_{y' \in \mathcal{O}(e)} u_{y', y}-n_{y, x} \leq M b_{x, e, y}  \label{mp:decep_lin_1} \\
                \sum_{y' \in \mathcal{O}(e)} u_{y', y}-n_{y, x} \geq -M b_{x, e, y}  \label{mp:decep_lin_1_2}
	        \end{align} 
	\end{enumerate}
\end{mdframed}%

     For each edge $e \in E$ and event $y \in Y $, we introduce an integer variable $c_{e, y}$, which is assigned the number of times the events in $\mathcal{O}(e)$ are mapped to $y$, through the following constraints:
    \begin{mdframed}
\small
	\begin{enumerate}[-]
	
		    

      \item $\forall e \in E$ and $y \in Y$, \vspace{-\baselineskip}
	        \begin{equation} \label{mp:decep_lin_3}
			   c_{e, y} = \sum_{y' \in \mathcal{O}(e)} u_{y', y}
		    \end{equation}  
      
	\end{enumerate}
\end{mdframed}%

    Additionally, for each word-observation $x \in \Sigma$ and edge $e \in E$, we introduce a binary variable $l_{x, e}$, which is assigned value $1$ iff $e$ produces $x$ by the sensor alteration attack.
    %
    %
    Using the big-M method for a sufficiently large $M'$, the following constraints are introduced to assign values to those variables.
\begin{mdframed}
\small
	\begin{enumerate}[-]
	\item $\forall x \in \Sigma$ and $e \in E$, \vspace{-1.8\baselineskip}
	        \begin{align} 
			   \sum_{y \in x} b_{x, e, y} + \sum_{y \in Y \setminus x} c_{e, y} \leq M' - M'l_{x, e} \label{mp:decep_lin_3_1} \\
               \sum_{y \in x} b_{x, e, y} + \sum_{y \in Y \setminus x} c_{e, y} \geq -M' + M'l_{x, e} \label{mp:decep_lin_3_2}
		    \end{align}  		          
	\end{enumerate}
\end{mdframed}%
    
    
    Finally, we use all this new introduced variables in the following constraints.
    \begin{mdframed}
\small
	\begin{enumerate}[-]
	\item $\forall q \in Q_{\mathrm{O}}$, $p \in Q_{\mathcal{M}}$, $x \in \Sigma$, and $e \in E$ s.t. $\delta_{\mathcal{M}}(p, e) \neq \bot$,
		    
	        \begin{equation} \label{mp:decep_lin_4}
      a_{q, p} \leq 1-l_{x, e} + a_{\delta_{\mathrm{O}}(q, x), \delta_{\mathcal{M}}(p, e)}
		    \end{equation}  		    
		    
	\end{enumerate}
\end{mdframed}%
%
%
The ILP formulation is now complete. We improve it in the following.  
Our ILP introduced a binary variable $l_{x, e}$ for each world-observation $x \in \Sigma$ and edge $e \in E$.
However, for a given environment, many of the elements within $\Sigma$ might not be realizable by the world-graph, meaning many of them might not be within $Z=\{ \mathcal{O}(e) \mid e \in E \}$.
Accordingly, for each state $q$ of the DFA $\mathrm{O}$, the state to which $q$ transitions by $x$ is an accepting state and from that accepting state, no non-accepting state of $O$ is reachable.
This is because the set of all world-observation sequences produced by the walks of the world-graph forms a prefix-closed language. In other words, for any world-observation sequence $t = z_1 z_2 \dots z \dots z_{n-1} z_{n}$ in which no $e \in E$ exists such that $z = \mathcal{O}(e)$, it holds $t \not\in L(\mathrm{P})$ or equivalently $t \in L(\mathrm{O})$. 
Recall that $\mathrm{P}$ accepts all the world-observation sequences produced for the walks in the itinerary and that $\mathrm{O}$ accepts all those sequences of world-observations that are either (1) not realizable by the world-graph or (2) are realizable by the world-graph but are generated for the walks outside the itinerary.
If $\mathrm{O}$ is minimized, then there is a unique sink state $q_{sink} \in F_{\mathrm{O}}$ s.t. for any $x \in Z$ and $q' \in Q_{\mathrm{O}}$, $\delta_{\mathrm{O}}(q', x) = q_{sink}$.
Note that it is possible that $|F_{\mathrm{O}}| \geq 1$, but such a unique state is guaranteed to exist if $\mathrm{O}$ is minimized. Other accepting state in  $F_{\mathrm{O}}$ are reached by the walks that are realizable by the world-graph but are not in the itinerary.

With these in mind, instead of defining variables $l_{x, e}$ for all $x \in \Sigma$ and $e \in E$, we define $l_{x, e}$ for only those $x$'s such that $x \in Z$.
Accordingly, all constraints of types (\ref{mp:decep_lin_1}), (\ref{mp:decep_lin_1_2}), (\ref{mp:decep_lin_3}), (\ref{mp:decep_lin_3_2}), and (\ref{mp:decep_lin_4}) are defined only for those $x \in Z$ rather than for all $x \in \Sigma$.
These may substantially reduce the number of variables and constraints.
However, we need more constraints to ensure the ILP is correct. 
If an edge $e$ does not produce any world-observation within $Z$ under the sensor attack, then
it means the world-observation produced for $e$ under the sensor attack reaches state $q_{sink}$ in $\mathrm{O}$.

With these in mind, we introduce the following constraints.
\begin{mdframed}
\small
	\begin{enumerate}[-]
	\item $\forall q \in Q_{\mathrm{O}}$, $p \in Q_\mathcal{M}$, and $e \in E$ s.t. $\delta_{\mathcal{M}}(p, e) \neq \bot$,
	        \begin{align} 
             a_{q, p} \leq a_{q_{sink}, \delta_{\mathcal{M}}(p, e)}+\sum_{x \in Z} l_{x, e}. \label{mp:decep_lin_5}
		    \end{align}  		          
	\end{enumerate}
\end{mdframed}%
By these constraints, if $a_{q, p} = 1$ and that $e$ does not produce any world-observation within $Z$ under the sensor attack, then it must hold $a_{q_{trap}, \delta_{\mathcal{M}}(p, e)}=1$, meaning that state $q$ enters the sink state $q_{sink}$ by the world-observation produced by $e$, regardless of what value that world-observation might have.

The time to construct the ILP is as follows.
\begin{lemma}
\label{lem:ilp_const_time}
   It takes $O(3|Y|^2+2|Z||E||Y|^2+2^{|Q_{\mathcal{I}}||V|+1}|V||Q_{\mathcal{D}}||Z||E|)$ in the worst case to construct the ILP.
\end{lemma}
See the supplementary for the proof.
\vspace{-10pt}
\section{Case Study}
\label{sec:case}
We present several instances of our problem, solved using our implementation of the algorithm in Python.
All executions were on a 2.80GHz Core i7 with a 16GB memory.

\begin{figure*}[t]

  \centering
  \includegraphics[width=1.0\linewidth]{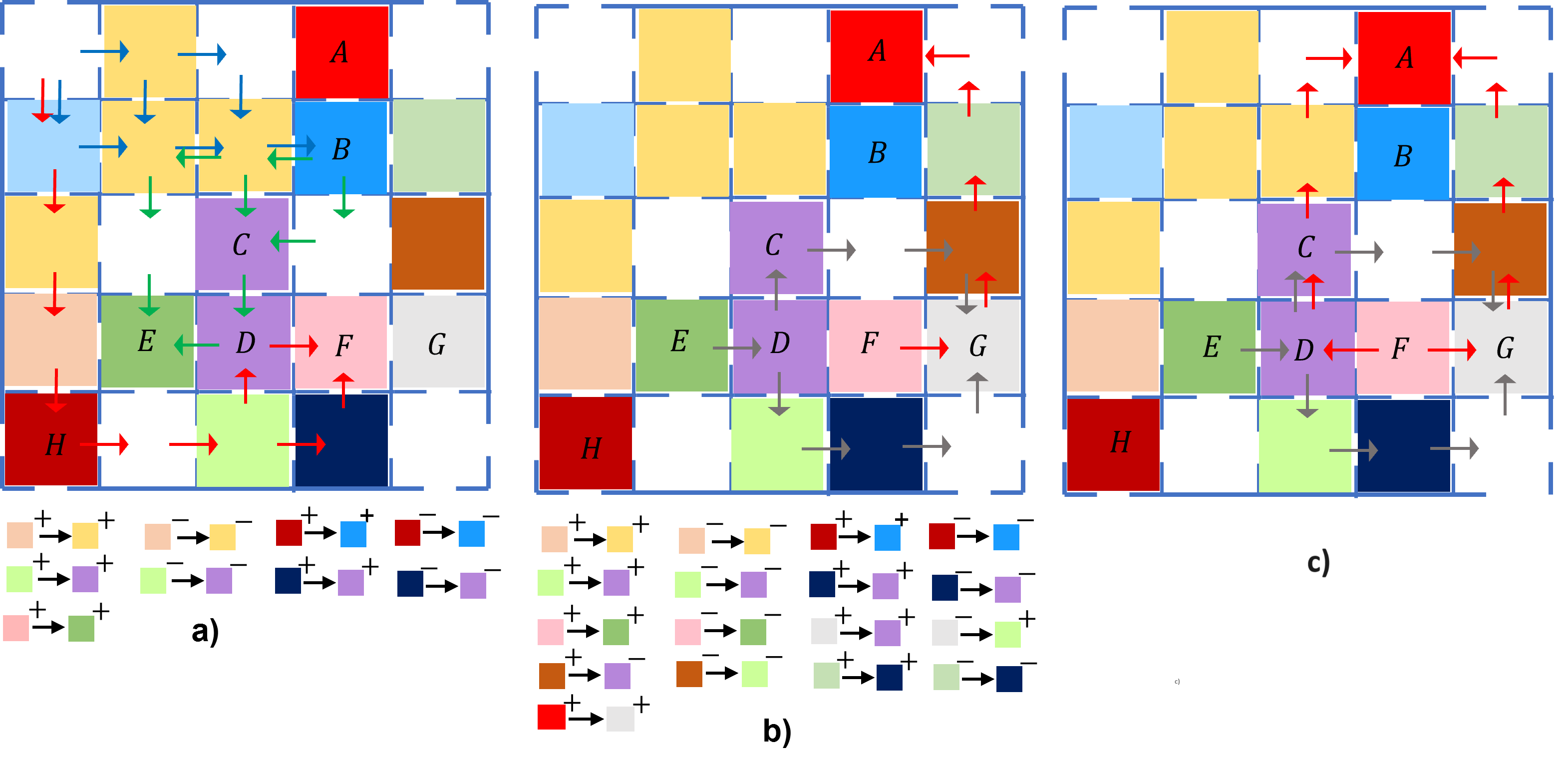}
  \vspace{-18pt}
  \caption{
     Sensor alteration planning in a gridworld.
     \textbf{a)} The itinerary: first visiting $B$ and then $E$, via blue and green arrows. The deviation: visiting $F$ via red arrows.
     \textbf{b)} The walks in Part a) are extended by the gray arrows for the itinerary and red arrows for the deviation.
     \textbf{c)} This instance also extends the walks in Part a). 
     Our algorithm indicates this instance has no feasible solution.
      }
      \label{fig:grid_5_5}
\vspace*{-18pt}
\end{figure*}

Fig.~\ref{fig:grid_5_5} shows the first instance.
Each room is guarded by an occupancy sensor or left unguarded.
Between each pair of adjacent rooms there is a door. 
%
%
There are $14$ sensors, producing a total of $28$ events, two events per sensor.
The agent starts at the top left corner, moves to $B$ via blue arrows, then to $E$ via green arrows, with the constraints of avoiding $A$ and $F$, and if entering $C$, the next room must be $D$. 
This yields $3$ allowed walks to $B$ and another $3$ from $B$ to $E$, totaling $9$ allowed walks.
%
%
%
The agent aims to visit first $H$ and then $F$, while avoiding $E$ and using only actions depicted by red arrows.
This deviation consists of two walks. 
%
The cost of altering each event to any other event was $1$.
The solution computed by our algorithm alters $9$ out of the $28$ existing sensor readings.
The solution computed by our implementation for this instance is shown in the bottom part of Fig.~\ref{fig:grid_5_5}a.
The cost of solution is $9$. It took $7.98$ seconds for our implementation to form the ILP and solving the ILP took $31.24$ seconds.
%

The next instance, shown in Fig.~\ref{fig:grid_5_5}b, extends each of the walks in the previous instance with $5$ actions, so that each walk in the itinerary can reach  $G$ using one of those two gray paths, while the walks in the deviation can reach $A$ using the red path.
The allow itinerary contains $9 \times 2= 18$ walks and the deviation contains $2 \times 1 = 2$ walks.
The solution computed by our implementation is shown in the figure, with a cost of $17$. It took $8.39$ seconds to form the ILP and $34.59$ seconds to solve it.
The third instance, shown in Fig.~\ref{fig:grid_5_5}c, extends the walks in the first instance by 
the two gray paths for the allowed itinerary and the two red paths for the deviation.
The allowed itinerary in this case contains $9 \times 2 = 18$ walks and the deviation contains $2 \times 2 = 4$ walks.
Forming the ILP for this case took $8.88$ seconds and it took $41.50$ seconds to find out the ILP has no feasible solution.
%
Recall that in a feasible solution, each walk within the deviation must be disguised as a walk within the itinerary.

The supplementary uploaded along with this manuscript presents our experiment results on $6$ additional instances of the problem on the environment in Fig.~\ref{fig:department}.
The execution time for each those instances is less than $1$ second.

\vspace{-\baselineskip}
\section{Conclusion} 
\label{sec:conclusion}
We formulated a sensor deception planning problem in which an adversary aims to optimally alter the sensor readings of a surveillance system to disguise its adversarial behaviors as allowed behaviors.
We proved the problem is $\NP$-hard and provided an integer linear programming algorithm for computing optimal and sub-optimal solutions.
We showed our algorithm is accurate and scalable via several experiments. The algorithm can be used to assess the vulnerabilities in the system's sensing capabilities, given possible knowledge about the attacker's intentions. 

We considered only the case where the sensor readings can be substituted but they cannot be suppressed.
Future work extends current work for situations where the agent has the option of suppressing some of the events, for example, by carrying out jamming attacks. 
 Future work can improve the ILP and consider standard steps dealing with $\NP$-hard problems, including providing heuristic and approximation algorithms. 

 \section*{Acknowledgments}
Research was sponsored by the Army Research Office and was accomplished under Grant
Number W911NF-22-1-0034 and Grant Number    W911NF-22-1-0166. The views and conclusions contained in this document are those of the authors and
should not be interpreted as representing the official policies, either expressed or implied, of the Army Research
Office or the U.S. Government. The U.S. Government is authorized to reproduce and distribute reprints for
Government purposes notwithstanding any copyright notation herein.

\bibliographystyle{splncs04}

\vspace{-10pt}
\bibliography{references}

\begin{thebibliography}{10}
\providecommand{\url}[1]{\texttt{#1}}
\providecommand{\urlprefix}{URL }
\providecommand{\doi}[1]{https://doi.org/#1}

\bibitem{fuAlmostSureIntentionDeception2022}
Fu, J.: On almost-sure intention deception planning that exploits imperfect observers. Springer International Publishing (Sep 2022)

\bibitem{hopcroft2001introduction}
Hopcroft, J.E., Motwani, R., Ullman, J.D.: Introduction to automata theory, languages, and computation. Acm Sigact News  \textbf{32}(1),  60--65 (2001)

\bibitem{karabagDeceptionSupervisoryControl2021a}
Karabag, M.O., Ornik, M., Topcu, U.: Deception in supervisory control. IRE Transactions on Automatic Control  (2021)

\bibitem{mastersDeceptivePathplanning2017}
Masters, P., Sardina, S.: Deceptive path-planning. In: Proc. {International} {Joint} {Conference} on {Artificial} {Intelligence}. pp. 4368--4375. Melbourne, Australia (Aug 2017)

\bibitem{meira2020synthesis}
Meira-G{\'o}es, R., Kang, E., Kwong, R.H., Lafortune, S.: Synthesis of sensor deception attacks at the supervisory layer of cyber--physical systems. Automatica  \textbf{121},  109172 (2020)

\bibitem{meira2019synthesis}
Meira-G{\'o}es, R., Kwong, R., Lafortune, S.: Synthesis of sensor deception attacks for systems modeled as probabilistic automata. In: Proc. American Control Conference. pp. 5620--5626. IEEE (2019)

\bibitem{meira2021synthesis}
Meira-G{\'o}es, R., Kwong, R.H., Lafortune, S.: Synthesis of optimal multiobjective attack strategies for controlled systems modeled by probabilistic automata. IEEE Transactions on Automatic Control  \textbf{67}(6),  2873--2888 (2021)

\bibitem{meira2021synthesissupervisors}
Meira-G{\'o}es, R., Lafortune, S., Marchand, H.: Synthesis of supervisors robust against sensor deception attacks. IEEE Transactions on Automatic Control  \textbf{66}(10),  4990--4997 (2021)

\bibitem{meyer1972equivalence}
Meyer, A.R., Stockmeyer, L.J.: The equivalence problem for regular expressions with squaring requires exponential space. In: SWAT. vol.~72, pp. 125--129 (1972)

\bibitem{mohajerani2020efficient}
Mohajerani, S., Meira-G{\'o}es, R., Lafortune, S.: Efficient synthesis of sensor deception attacks using observation equivalence-based abstraction. IFAC-PapersOnLine  \textbf{53}(4),  28--34 (2020)

\bibitem{phatak2023sensor}
Phatak, R., Shell, D.A.: Sensor selection for fine-grained behavior verification that respects privacy. In: 2023 IEEE/RSJ International Conference on Intelligent Robots and Systems (IROS). pp. 8628--8635. IEEE (2023)

\bibitem{rahmani2021sensor}
Rahmani, H., Shell, D.A., O’Kane, J.M.: Sensor selection for detecting deviations from a planned itinerary. In: Proc. IEEE/RSJ International Conference on Intelligent Robots and Systems. pp. 6511--6518. IEEE (2021)

\bibitem{wangSupervisoryControlDiscrete2019}
Wang, Y., Pajic, M.: Supervisory control of discrete event systems in the presence of sensor and actuator attacks. In: Proc. {IEEE} {Conference} on {Decision} and {Control}. pp. 5350--5355 (Dec 2019), iSSN: 2576-2370

\bibitem{yao2022sensor}
Yao, J., Yin, X., Li, S.: Sensor deception attacks against initial-state privacy in supervisory control systems. In: Proc. IEEE Conference on Decision and Control. pp. 4839--4845. IEEE (2022)

\bibitem{yu2010cyber}
Yu, J., LaValle, S.M.: Cyber detectives: Determining when robots or people misbehave. In: WAFR. pp. 391--407. Springer (2010)

\bibitem{yu2011story}
Yu, J., LaValle, S.M.: Story validation and approximate path inference with a sparse network of heterogeneous sensors. In: 2011 IEEE International Conference on Robotics and Automation. pp. 4980--4985. IEEE (2011)

\bibitem{zheng2021modeling}
Zheng, S., Shu, S., Lin, F.: Modeling and control of discrete event systems under joint sensor-actuator cyber attacks. In: Proc. International Conference on Automation, Control and Robotics Engineering. pp. 216--220. IEEE (2021)

\end{thebibliography}

\cleardoublepage
\begingroup
\let\clearpage\relax
\title{Supplementary}
\author{Hazhar Rahmani\inst{1}\orcidID{0000-1111-2222-3333} \and
Second Author\inst{Arash Ahadi}\orcidID{1111-2222-3333-4444} \and
Jie Fu\inst{3}\orcidID{2222--3333-4444-5555}}
\authorrunning{H. Rahmani et al.}
%
%
%

%
%
%
%
%
%
\section{Supplementary}
\subsection{Computational Complexity of the Algorithm}
%
\textbf{Proof of Lemma 3:}
\begin{proof}
The improved ILP has $|Q_{\mathrm{O}}||Q_{\mathcal{M}}|$ variables for the $a_{q, p}$'s, $|Y|^2$ variables for the $u_{y, y'}$'s, $|Y||Z|$ variables for the $n_{y, x}$'s, $|Z||E||Y|$ variables for the $b_{x, e, y}$'s, $|E||Y|$ variables for the $c_{e, y}$'s, and $|Z||E|$ variables for the $l_{x, y}$'s, where $|Z| = 2^{|\{ \mathcal{O}(e) \mid e \in E \}|}$. 
Note that because $\mathcal{M}$ is the product of $\mathcal{D}$ and $\mathcal{G}$, $|Q_{\mathcal{M}}|=O(|V||Q_{\mathcal{D}}|)$. Also, given that $\mathrm{O}$ is made from $\mathrm{P}$ using the powerset construction of NFA to DFA conversion  and that $|Q_{\mathrm{P}}|=O(|V||Q_{\mathcal{I}}|)$ (because $\mathrm{P}$ is made from $\mathcal{P}$ and that $\mathcal{P}$ is the product of $\mathcal{G}$ and $\mathcal{I}$), $|Q_{\mathrm{O}}|=O(2^{|Q_{\mathcal{I}}||V|})$. Note that this gives a worst case analysis because the powerset construction takes an exponential time in the worst case, but for practical itinerary DFAs, the powerset construction might be performed efficiently.
Therefore, the total number of variables is 
$O(2^{|Q_{\mathcal{I}}||V|}|V||Q_{\mathcal{D}}|+|Y|(|Y|+|Z|+|Z||E|+|E|)+|Z||E|)$.
The ILP has 1 constraint of type (\ref{mp:init}) with size $O(1)$, $|Y|$ constraints of type (\ref{mp:mapping}) with size $O(|Y|)$, $O(|Y|^2)$ constraints of type (\ref{mp:infeasible_mapping}) with size $O(1)$, $|F_{\mathrm{O}}||F_{\mathcal{M}}|$ constraints of type (\ref{mp:deceptive}) with size $O(1)$, $|Q_{\mathrm{O}}||Q_{\mathcal{M}}|$ constraints of type (\ref{mp:a}) with size $O(1)$, $|Y|^2$ constraints of type (\ref{mp:u}) with size $O(1)$, $|Z||E||Y|$ constraints of each types (\ref{mp:decep_lin_1}) and (\ref{mp:decep_lin_1_2}) each with size $O(|Y|)$, $|E||Y|$ constraints of type (\ref{mp:decep_lin_3}) with size $O(|Y|)$, $|Z||E|$ constraints of each types (\ref{mp:decep_lin_3_1}) and (\ref{mp:decep_lin_3_2}) each with size $O(|Y|)$, $O(|Q_{\mathrm{O}}||Q_{\mathcal{M}}||Z||E|)$ constraints of type (\ref{mp:decep_lin_4}) with size $O(1)$, and $O(|Q_{\mathrm{O}}||Q_{\mathcal{M}}||E|)$ constraints of type (\ref{mp:decep_lin_5}) with size $O(|Z|)$.
Therefore, the ILP has $1+|Y|+O(|Y|^2)+|F_{\mathrm{O}}||F_{\mathcal{M}}|+|Q_{\mathrm{O}}||Q_{\mathcal{M}}|+|Y|^2+2|Z||E||Y|+|E||Y|+2|Z||E|+O(|Q_{\mathrm{O}}||Q_{\mathcal{M}}||Z||E|)+O(|Q_{\mathrm{O}}||Q_{\mathcal{M}}||E|)$ constraints.
The running time to construct the ILP using appropriate data structures, mainly hash maps, is $O(1+3|Y|^2+|F_{\mathrm{O}}||F_{\mathcal{M}}|+|Q_{\mathrm{O}}||Q_{\mathcal{M}}|+2|Z||E||Y|^2+|E||Y|^2+2|Z||E||Y|+2|Q_{\mathrm{O}}||Q_{\mathcal{M}}||Z||E|)$.
Simplifying this and replacing $|Q_{\mathrm{O}}|$ and $|Q_{\mathcal{M}}|$ yields the following running time for constructing the ILP: $O(3|Y|^2+2|Z||E||Y|^2+2^{|Q_{\mathcal{I}}||V|+1}|V||Q_{\mathcal{D}}||Z||E|)$.
\end{proof}

\subsection{Baseline algorithm}
In section, we discuss a baseline algorithm to solve \MCSD.
This algorithm exhausts all the sensor alterations $A: Y \rightarrow Y$, perhaps from the one with the lowest cost to the one with the highest cost, and for each sensor alteration $A$ uses Definition~\ref{def:nfas} to construct NFAs $\mathrm{P}$ and $\mathrm{M}$. Then, it uses the result in Corollary~\ref{cor:decept_sensor} to check if $A$ is deceptive or not. 
Then, among all those sensor alterations that are deceptive, it chooses the one that has the minimum cost.
Because $\mathrm{P}=(Q_{\mathrm{P}}, \Sigma, \delta_{\mathrm{P}}, q_0^{\mathrm{P}}, F_{\mathrm{P}})$ and $\mathrm{M}=(Q_{\mathrm{M}}, \Sigma, \delta_{\mathrm{M}}, q_0^{\mathrm{M}}, F_{\mathrm{M}})$  are NFAs, to use the result in Corollary~\ref{cor:decept_sensor}, one first needs to convert those two NFAs into DFAs, and check if the intersection of the languages of those two DFAs is the empty string or not.
Let $\mathrm{B}=(Q_{\mathrm{B}}, \Sigma, \delta_{\mathrm{B}}, q_0^{\mathrm{B}}, F_{\mathrm{B}})$ and 
$\mathrm{C}=(Q_{\mathrm{C}}, \Sigma, \delta_{\mathrm{C}}, q_0^{\mathrm{C}}, F_{\mathrm{C}})$ be  two DFAs that are respectively equivalent to $\mathrm{P}$ and $\mathrm{M}$.
Sensor selection $A$ is deceptive if and only if the DFA $K=(Q_{\mathrm{B}} \times Q_{\mathrm{C}}, \Sigma, \delta_K, (q_0^{\mathrm{B}}, q_0^{\mathrm{C}}),  F_{\mathrm{B}} \times  F_{\mathrm{C}} ))$ in which $\delta_K((q, p), a) = (\delta_{\mathrm{B}}(q, a), \delta_{\mathrm{C}}(p, a))$ for each $(q, p) \in Q_{\mathrm{B}} \times Q_{\mathrm{C}}$ and $a \in \Sigma$, has no reachable accepting state. 
The running time of this algorithm in the worst case is $O(|Y|!(2^{|V| \times |Q_{\mathcal{I}}|} 2 ^ {|V| \times |Q_{\mathcal{D}}|} |\Sigma| ))$.
%
Our algorithm and this baseline algorithm are the only known algorithms to solve \MCSD.

\subsection{Case Study 2: Sensor alteration   in a small   environment}
We executed several instances of \MCSD, with different itinerary DFAs and different deviation DFAs, using the world-graph in Fig.~\ref{fig:department} to verify correctness of our algorithm.
Table~\ref{tab:caseStudy1} shows results of this experiment.
This table shows for each instance, the time to form the ILP (the left term) and the time to solve the ILP (the right term).
For convenience, all the itineraries and deviations are expressed using regular expressions. 
In all those scenarios, $c(y, y')=1$ for each $y, y' \in Y$ such that $y \neq y'$, and $c(y, y) = 0$ for each $y \in Y$, meaning that the cost of altering any event to any other event is 1.
The first two instances are extreme boundary test cases. In both, the itinerary DFA consists of all walks in the word-graph, including $\epsilon$, which represents not taking any transition. 
For each of which, our implementation of the algorithm indicates there is no need to alter any sensor.
In the first case, the deviation consists of not taking any transition, and in the second case, the deviation consists of all walks in the graph, including $\epsilon$.
Trivially, for each of these two cases, there is no need to alter any sensor reading.
In the third instance, each of the itinerary and the deviation consists of a single, simple walk, for which our algorithm alters $b_3$ to $b_4$ and $o_1^+$ to $o_3^+$.
The itinerary in the fourth consists of all the walks where, after transitioning between the regions in the corridor as many times as it wants, the agent enters either rooms $F$ and $G$ and then exits after an arbitrary number of transitions between rooms $F$ and $G$.
Note that this itinerary consists of infinite number of walks.
The deviation consists of a single walk.
Our implementation alters $b_1$ to $b_3$ and maps both $b_2$ and $b_4$ to $b_5$.
In fact, in this case, the single walk within the deviation is mapped to the walk $e_{17} e_{21} e_{22} e_{18}$.
For the allowed itinerary in the fifth row, the agent is allowed to enter room $G$, then enter room $F$ via either the left door or the right door, and then exist room $F$. Accordingly, the allowed itinerary consists of two walks. 
The agent intends to enter $C$ and from there enter $D$ and then exit $D$.
Our algorithm indicates there is no deceptive sensor attack for this case. 
This is because in the last transition of the walk within the deviation only one sensor event occurs, but in the last transition of any of the two walks within the itinerary, two events $b_6$ and $o_3^-$ occur simultaneously and the agent is not allowed to suppress an event.
The itinerary in the last instance consists of all walks in which the agent enters room $F$ or $G$ with possible transitions between them and then either visits room $C$ or room $A$ and then exits.
The deviation consists of two walks, in both of which the agent passes through rooms $C$ and $D$, and then it either visits room $E$ and exits or visit room $B$ and exits.
The sensor alteration computed by our algorithm alters $9$ sensor readings.

\begin{table*}
            \centering    
            \vspace*{-10pt}
\begin{tabular}{c @{\hspace{2mm}} l l l l}
    \hline
        & \textbf{Itinerary} & \textbf{Deviation} & \textbf{Computed solution} & \textbf{Comp.} \\ 
        &  &  &  & \textbf{time (sec)} \\ 
       \hline
        1 & $(e_1 | e_2 | \cdots | e_{26})^*$ & $\epsilon$ & Every event is &  \\ 

         & Any walk, including $\epsilon$,  & Not taking any transition &  mapped to itself & 0.39+0.31 \\ 

         & not taking any transition &  &   &  \\ 
         
        \hline

         2 & $(e_1 | e_2 | \cdots | e_{26})^*$ & $(e_1 | e_2 | \cdots | e_{26})^*$ & Every event is &  \\ 

         & Any walk, including $\epsilon$, & Any walk, including $\epsilon$,  &  mapped to itself & 0.32+0.28 \\ 

        &  not taking any transition & not taking any transition &   &  \\ 
         
        \hline
       
        3 & $e_9 e_{25}$ & $e_1 e_3$ &  &  \\ 

         & Directly go to room $E$ & Directly go to room $A$ & $b_3 \rightarrow b_4, o_1^+ \rightarrow o_3^+$ & 0.20+0.26 \\ 
        
        \hline
        4 & $(e_1|e_2|e_9|e_{10})^*(e_{17}|e_{19})\cdots$ &  &  &   \\

         & $(e_{21}|e_{22}|e_{23}|e_{24})^*(e_{18}|e_{20})$ &  &  &   \\
        
         & Visit rooms $F$ or $G$, with  & Take a clockwise tour  & $b_1 \rightarrow b_3, b_2 \rightarrow b_5, $ & 
        \\ 
        & possible transitions between &  the corridor and room $B$,  &  & 
        \\ 
        
         & them & without visiting a region twice  & $b_4 \rightarrow b_5$ & 0.32+0.30
        \\ 
        \hline

        5 & $e_{17} (e_{21}|e_{23}) e_{20}$ & $e_7 e_{11} e_{15}$ & & \\
          & Visit rooms $G$ or $F$, with   & Go to room $C$, then  & \\
          &  possible transitions between  &  to room $D$, then to & \\
          & them, then go to the corridor &  the corridor & \textsc{Infeasible} & 0.29+0.20 \\
          \hline

        6 & $(e_{17} | e_{19})(e_{21} | e_{22} | e_{23} | e_{24})^*$ & $e_7 e_{11} e_{15} (e_{14} e_5 e_2 | e_{25} e_{26} e_{10})$ & $b_1 \rightarrow b_3, b_2 \rightarrow b_5$ & \\
         & $(e_{18} | e_{20})(e_1 e_7 e_3 e_4)^*(e_2 | e_8)$ & $e_7 e_{11} e_{15} (e_{14} e_5 e_2 | e_{25} e_{26} e_{10})$ & $b_1 \rightarrow b_3, b_2 \rightarrow b_5$ & \\
          & Visit rooms $G$ or $F$, with   & Go to room $C$, then to  & $b_4 \rightarrow b_3, o_1^+ \rightarrow b_5$ & \\
          & possible transitions between  &  room $D$, then either & $b_4 \rightarrow b_3, o_1^+ \rightarrow b_5$ & \\
          & them, then either visit room D  &  visit room $B$ and exit or & $o_1^- \rightarrow o_3^-, o_2^+ \rightarrow b_5$ &  \\
          & and exit or visit A as many &  or visit room $E$ and exit  & $o_1^- \rightarrow o_3^-, o_2^+ \rightarrow b_5$ &  \\
          & times as it wanted and exit &  & $o_2^- \rightarrow o_3^+, o_3^+ \rightarrow b_5$ &  \\
           &  &  & $o_3^- \rightarrow b_3$ & 0.46+0.34 \\
        
    \hline
\end{tabular}

    \caption{Results of our implementation for the example in Fig.~\ref{fig:department}. A computation time consists of the time to form the ILP and the time to solve the ILP.}
            \label{tab:caseStudy1}
\end{table*}


\endgroup

\end{document}